\documentclass[journal,compsoc,twoside]{IEEEtran}

\usepackage{graphicx}
\usepackage{epsfig}
\usepackage{amsmath}
\usepackage{amssymb}
\usepackage{textcomp}
\usepackage{bbm}
\usepackage[american]{babel}

\usepackage{balance}
\usepackage[colorlinks, citecolor=blue, bookmarks=true, CJKbookmarks=true, bookmarksnumbered=true,bookmarksopen=true]{hyperref}

\usepackage[T1]{fontenc}    %
\usepackage{microtype}      %

\makeatletter
\newif\if@restonecol
\makeatother

\usepackage[linesnumbered,ruled,vlined]{algorithm2e}%
\usepackage{algpseudocode}
\usepackage{amsmath}

\usepackage{bm}%

\usepackage{multirow} 

\newcommand{\squishlist}{
	\begin{list}{$\bullet$}
		{ \setlength{\itemsep}{0pt}
			\setlength{\parsep}{2pt}
			\setlength{\topsep}{2pt}
			\setlength{\partopsep}{0pt}
			\setlength{\leftmargin}{1em}
			\setlength{\labelwidth}{1em}
			\setlength{\labelsep}{0.5em} } }
	\newcommand{\squishend}{
\end{list} }

\makeatletter
\DeclareRobustCommand\onedot{\futurelet\@let@token\@onedot}
\def\@onedot{\ifx\@let@token.\else.\null\fi\xspace}

\def\etal{\emph{et al}\onedot}
\makeatother

\usepackage{amsthm}
\newtheorem{definition}{Definition}%
\newtheorem{theorem}{Theorem}%
\newtheorem{corollary}{Corollary}%
\newtheorem{lemma}{Lemma}%
\newtheorem{proposition}{Proposition}%

\begin{document}

	\title{From Open Set to Closed Set: Supervised Spatial Divide-and-Conquer for Object Counting}

	\author{Haipeng Xiong, Hao Lu, Chengxin Liu, Liang Liu, Chunhua Shen, Zhiguo Cao\\
		\thanks{
			This work is supported by the Natural Science Foundation of China under Grant No.\  61876211. \textit{(Corresponding author: Zhiguo Cao. 
			H. Xiong and H. Lu contributed equally.
			Part of the work was done when H. Xiong was visiting The University of Adelaide.%
			)}
			
			H. Xiong, C. Liu, L. Liu, Z. Cao are with the National Key Laboratory of Science and Technology on Multi-Spectral Information Processing, School of Artificial Intelligence and Automation, Huazhong University of Science and Technology, Wuhan 430074, China (e-mail: hpxiong@hust.edu.cn, cx\_liu@hust.edu.cn, wings@hust.edu.cn, zgcao@hust.edu.cn).
			
			H. Lu and C. Shen 
			are with
			the University of Adelaide, SA 5005, Australia (e-mail: hao.lu@adelaide.edu.au, chunhua.shen@adelaide.edu.au).
		}%
	}

	\IEEEtitleabstractindextext{%
		\begin{abstract}
			
			Visual counting, a task that aims to estimate the number of objects from an image/video, is an open-set problem by nature
			as 
			the number of population can vary in $[0,+\infty)$ in theory. However, collected data are limited in reality, which means that only a closed set is observed. Existing methods typically model this task through regression, while they are prone to suffer from unseen scenes with counts out of the scope of the closed set. In fact, counting has an interesting and exclusive property---spatially decomposable. A dense region can always be divided until sub-region counts are within the previously observed closed set. We therefore introduce the idea of spatial divide-and-conquer (S-DC) that transforms open-set counting into a closed set problem. This idea is implemented by a novel Supervised Spatial Divide-and-Conquer Network (SS-DCNet). 
		    It can learn from a closed set but generalize to open-set scenarios via S-DC. 
			We provide theoretical analyses and a controlled experiment on
			synthetic 
			data, demonstrating why closed-set modeling 
			works well.
			Experiments show that SS-DCNet achieves state-of-the-art performance in crowd counting, 
			vehicle counting and plant counting.
			SS-DCNet also demonstrates superior transferablity under the cross-dataset setting.
			Code and models 
			are 
			available at: \url{https://git.io/SS-DCNet}.
			
		\end{abstract}
		
	}
	
	\maketitle
	\section{Introduction}\label{sec:intro}

\begin{figure}[t]
		\begin{center}
			\includegraphics[width=1.0\linewidth]{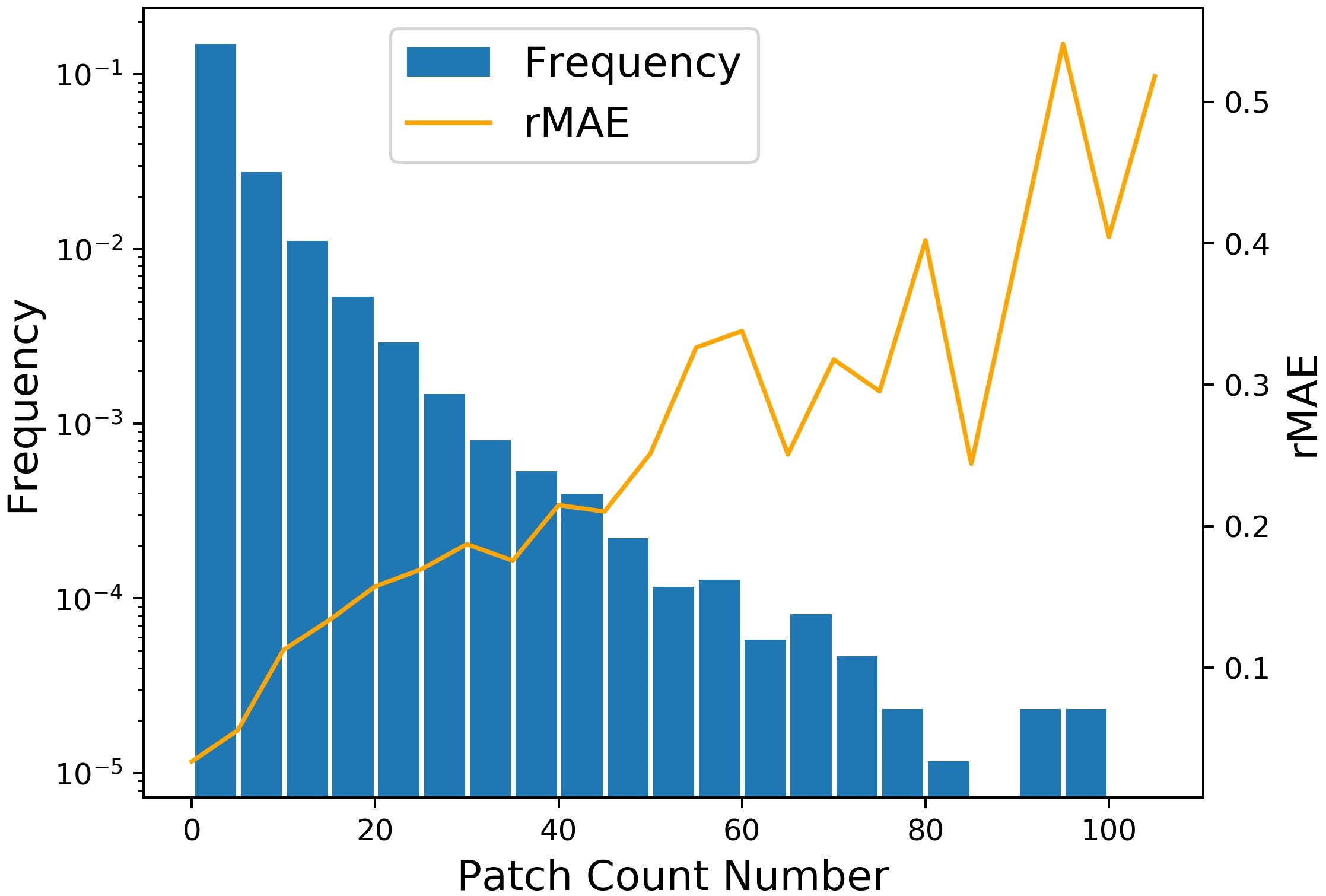}
		\end{center}
		\vspace{-10pt}
		\caption{The histogram of count values of $64\times64$ local patches on the test set of ShanghaiTech Part\_A dataset~\cite{MCNN_2016_CVPR}. The orange curve denotes the relative mean absolute error (rMAE) of CSRNet~\cite{CSRNet_2018_CVPR} on local patches. }
		\label{fig:image_dis_rmae}
	\end{figure}

	Counting is an open-set problem by nature as a count value can range from $0$ to $+\infty$ in theory. It is therefore typically modeled in a regression manner. Benefiting from the success of convolutional neural networks (CNNs), state-of-the-art deep counting networks often adopt a multi-branch architecture to enhance the feature robustness 
	 to 
	dense regions in an image~\cite{SwitchCNN_2017_CVPR,SANet_2018_ECCV,MCNN_2016_CVPR}. However, the observed patterns in datasets are limited in practice, which means that networks can only learn from a \textit{closed} set. Are these counting networks still able to 
	produce 
	accurate predictions when \emph{the number of objects is out of the scope of the closed set?}
	According to Fig.~\ref{fig:image_dis_rmae}, local counts observed in the closed set exhibit a long-tailed distribution. Extremely dense patches are rare while sparse patches take up the majority. As what can be observed, increased local density leads to significantly 
	deteriorated performance in 
	relative mean absolute error (rMAE). \emph{Is it necessary to set the working range of CNN-based counters to the maximum count value observed, even 
	though 
	a majority of samples are sparse %
	and 
	the counter works poorly in this range?}
	
	\begin{figure}[t]
		\begin{center}
			\includegraphics[width=1.0\linewidth]{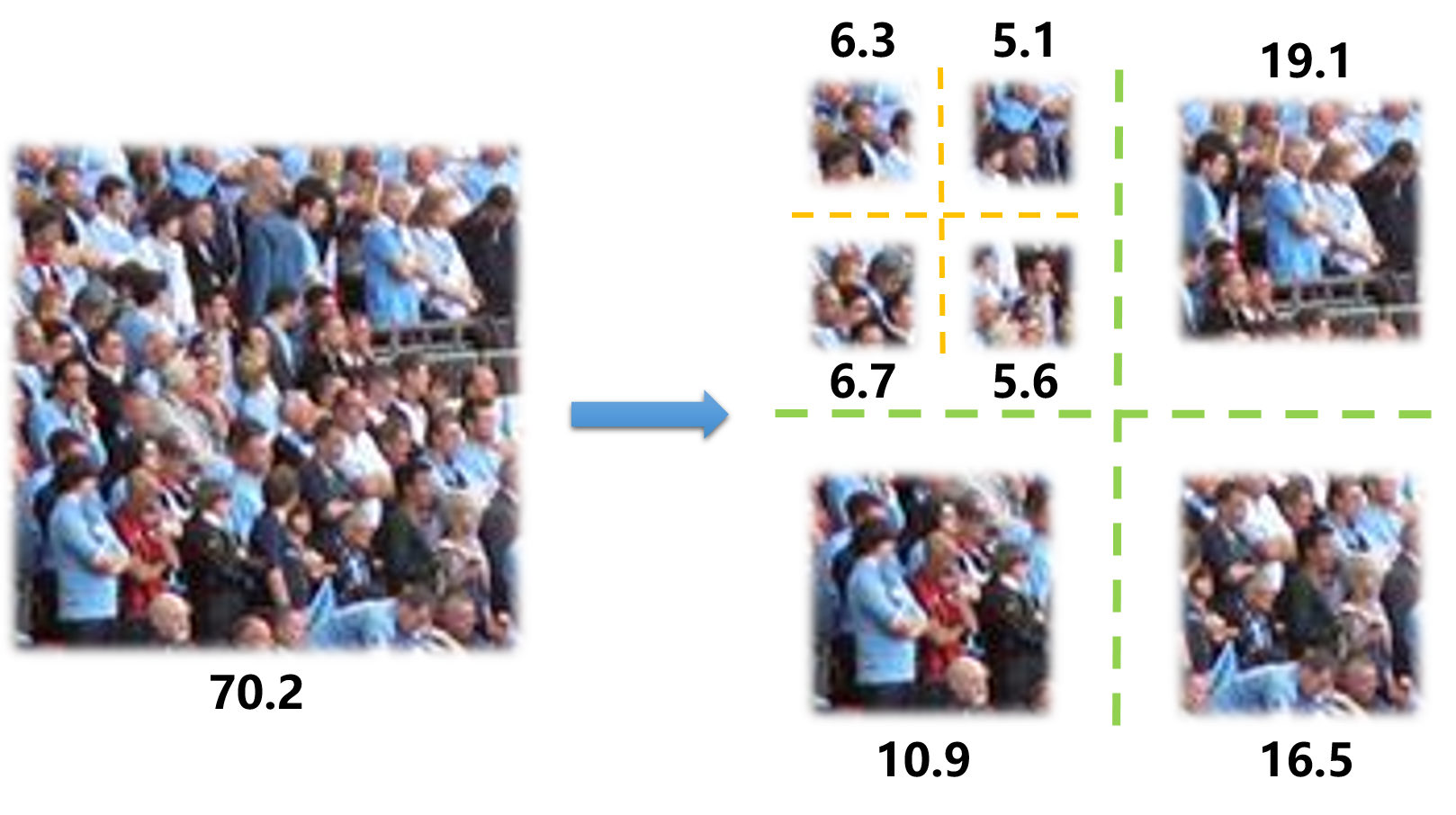}
		\end{center}
		\vspace{-10pt}
		\caption{An illustration of spatial divisions. Suppose that the closed set of counts is $[0, 20]$. In this example, dividing the image for one time is inadequate to ensure that all sub-region counts are within the closed set. For the top left sub-region, it needs a further division. }
		\label{fig:image_divide_example}
	\end{figure}
	
\begin{figure}[t]
		\begin{center}
			\includegraphics[width=1.0\linewidth]{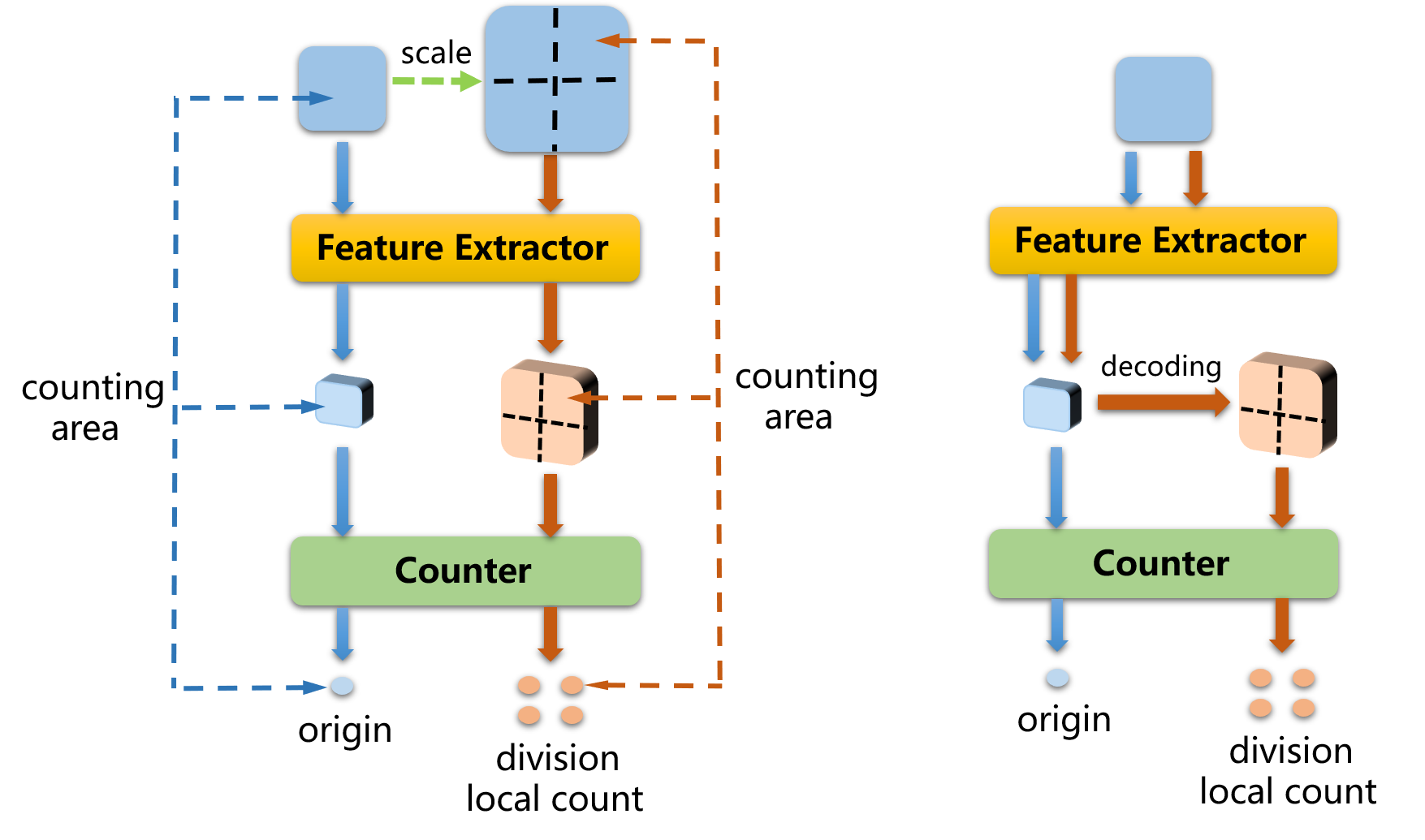}
		\end{center}
		\vspace{-10pt}
		\caption{Spatial divisions on the input image (left) and the feature map (right). Spatially dividing the input image is straightforward. The image is upsampled and fed to the same network to infer counts of local areas. The orange dashed line is used to connect the local feature map, the local count and the sub-image. S-DC on the feature map avoids redundant computations and is achieved by upsampling, decoding and dividing the feature map of high resolution.}
		\label{fig:feature_divide}
	\end{figure}

	In fact, counting has an interesting and exclusive property---being spatially decomposable. The above problem can be largely alleviated with the idea of spatial divide-and-conquer (S-DC). Suppose that a network has been trained to accurately predict a closed set of counts, say $0\sim20$. When facing an image with extremely dense objects, one can keep dividing the image into sub-images until all sub-region counts are less than $20$. The network can then count these sub-images and sum over all local counts to obtain the global image count. Fig.~\ref{fig:image_divide_example} %
	depicts the overall idea of S-DC. %
	A 
    subsequent 
	question is how to spatially divide the count. A naive
	approach 
	is to upsample the input image, divide it into sub-images and process sub-images with the same network. This %
	approach, 
	however, is likely to blur the image and lead to exponentially-increased %
	computation cost and memory consumption. Inspired by fully convolutional networks and 
	RoI pooling~\cite{girshick2015fast}, we show that it is feasible to achieve S-DC on
	feature maps, as
	shown 
	in Fig.~\ref{fig:feature_divide}. By decoding and upsampling the feature map, the 
	following
	prediction layers can focus on the feature of local areas and predict sub-region counts accordingly.
	
	To implement the idea above, we propose a simple 
	yet effective Supervised Spatial Divide-and-Conquer Network (SS-DCNet). SS-DCNet learns from a closed set of count values but is able to generalize to open-set scenarios. Specifically, \mbox{SS-DCNet} adopts a VGG16~\cite{Simonyan2014Very_VGG16}-based encoder and an UNet~\cite{Unet2015U}-like decoder to generate multi-resolution feature maps. All feature maps share the same counter. The counter can be designed by following the standard local count regression paradigm~\cite{Lu2017TasselNet} or by discretizing continuous count values into a set of intervals as a classifier following~\cite{Li2018depth,liu2019classification}. Furthermore, a division 
	decision module 
	is designed to decide which sub-region should be divided and to merge different levels of sub-region counts into the global image count.

	We provide theoretical analyses to shed light on why the transition from the open set to the closed set makes sense for counting. We also show through a controlled 
	experiment on synthetic data that, even given a closed training set, SS-DCNet effectively generalizes to the open test set. The effectiveness of \mbox{SS-DCNet} is further demonstrated on three crowd counting datasets (ShanghaiTech~\cite{MCNN_2016_CVPR}, UCF\_CC\_50~\cite{UCFCC50_2013_CVPR} and UCF-QNRF~\cite{Compose_Loss_2018_ECCV}), a vehicle counting dataset (TRANCOS~\cite{TRANCOSdataset_IbPRIA2015}), and a plant counting dataset (MTC~\cite{Lu2017TasselNet}). Results show that SS-DCNet indicates a clear advantage over other competitors and sets the new state of the art. In addition, we remark that the closed set of SS-DCNet executes an implicit transfer in the output space, which is backed by state-of-the-art performance under the cross-domain evaluations. In particular, SS-DCNet even beats most state-of-the-art counting models that are trained directly on the target domain in the task from UCF-QNRF to ShanghaiTech Part\_A.
	
	In summary, 
	the main contributions of this work are as follows.
	\squishlist
	\item  We propose to transform open-set counting into a closed-set problem via S-DC. A theoretical analysis of why such a transformation 
	works well 
	is also presented; 
	\item We investigate the explicit supervision for S-DC, which leads to a novel SS-DCNet. SS-DCNet is applicable to both regression-based and classification-based counters and can produce visually clear spatial divisions;
	
	\item We report state-of-the-art counting performance over $5$ challenging datasets with remarkable relative improvements. 
	We also show  good transferablity of SS-DCNet via cross-dataset evaluations on crowd counting datasets.
	\squishend

	A preliminary conference version of this work appeared in~\cite{xhp2019SDCNet} where S-DCNet, the first version of SS-DCNet, was developed. Here we have  extended~\cite{xhp2019SDCNet} in the following aspects: i) we provide theoretical analyses why closed set modeling 
	works well;
	ii) we further enhance S-DCNet at the methodology level by investigating further a regression-based closed-set counter, by integrating a count-orientated upsampling operator and by improving the model training with explicit supervision of spatial divisions; iii) we provide more 
	ablative studies and qualitative analyses to highlight the role of S-DC; and iv) we give an insight of SS-DCNet w.r.t.\  its good transferablity in the output space and report state-of-the-art performance under the %
	cross-dataset evaluation 
	setup.

	\section{Related Work}
	Current CNN-based counting approaches are mainly built upon the framework of local regression. According to their regression targets, they can be categorized into two categories: density map regression and local count regression. We first review these two regression paradigms. Since SS-DCNet works not only in regression counts but also in classification, some works that reformulate the regression problem are also discussed.
	
	\subsection{Density Map Regression}
	The concept of density map was introduced in~\cite{vlaz2010denlearn}. The density map contains the spatial distribution of objects, thus can be smoothly regressed. Zhang~\etal~\cite{Zhang_2015_CVPR} may be the first to adopt a CNN to regress local density maps. Then almost all subsequent counting networks followed this idea. Among them, a typical network architecture is multi-branch. MCNN~\cite{MCNN_2016_CVPR} and Switching-CNN~\cite{SwitchCNN_2017_CVPR} used three columns of CNNs with varying receptive fields to depict objects of different scales. SANet~\cite{SANet_2018_ECCV} adopted Inception~\cite{GoogleNet_2015_CVPR}-liked modules to integrate extra branches. 
	CP-CNN~\cite{CPCNN_2017_ICCV} added two extra density-level prediction branches to combine global and local contextual information. ACSCP~\cite{ACSCP_2018_CVPR} inserted a child branch to match cross-scale consistency and an adversarial branch to attenuate the blurring effect of the density map. ic-CNN~\cite{ICNN_2018_ECCV} incorporated two branches to generate high-quality density maps in a coarse-to-fine manner. IG-CNN~\cite{Divide_grow_2018_CVPR} and D-ConvNet~\cite{DeepNegCor_2018_CVPR} drew inspirations from ensemble learning and trained a series of networks or regressors to tackle different scenes. DecideNet~\cite{DecideNet_2018_CVPR} attempted to selectively fuse the results of density map estimation and object detection for different scenes. Unlike multi-branch approaches, Idrees~\etal~\cite{Compose_Loss_2018_ECCV} employed a composition loss and simultaneously solved several counting-related tasks to assist counting. CSRNet~\cite{CSRNet_2018_CVPR} benefited from dilated convolution which effectively expanded the receptive field to capture contextual information.
	
	Existing deep counting networks aim to generate high-quality density maps. However, density maps are actually in the open set as well. For a single point, different kernel sizes lead to different density values. When multiple objects exist and are close, density patterns are even much diverse. Since observed samples are limited, density maps are 
	clearly 
	in an open set. In addition, density maps do not have the physical property of spatial decomposition. We therefore cannot apply S-DC to density maps.

	\subsection{Local Count Regression}
	Local count regression directly predicts count values of local image patches. This idea first appeared in~\cite{chen2012feature} where a multi-output regression model was used to regress region-wise local counts simultaneously. Authors of \cite{Count_ception_2017_ICCVW} and~\cite{Lu2017TasselNet} introduced such an idea into deep counting. Local patches were first densely sampled in a sliding-window manner with overlaps, and a local count was then assigned to each patch by the network. Inferred redundant local counts were finally normalized and fused to the global count. Stahl~\etal~\cite{tip2019divide} regressed the counts for object proposals generated by Selective Search~\cite{Uijlings2013Selective} and combined local counts using an inclusion-exclusion principle. Inspired by subitizing, the ability for a human to quickly counting a few objects at a glance, Chattopadhyay~\etal~\cite{Count_everyday_2017_CVPR} transferred their focus to the problem of counting objects in everyday scenes. The main challenge thus shifted to large intra-class variances rather than the occlusions and perspective distortions in crowded scenes.

	While some methods above~\cite{Count_everyday_2017_CVPR,tip2019divide} leverage the idea of spatial divisions, they still regress the open-set counts. Despite the fact that local region patterns are easier to be modelled than the whole image, the observed local patches are still limited. Since only finite local patterns (a closed set) can be observed, new scenes in reality have a high probability including objects out of the range (an open set). Moreover, dense regions with large count values are rare (Fig.~\ref{fig:image_dis_rmae}) and the networks may suffer from sample imbalance. In this paper, we show that \textit{a counting network is able to learn from a closed set with a certain range of counts, e.g., $0\sim20$, and then generalizes to an open set (including counts $>20$) via S-DC}. 

	\subsection{Beyond Simple Regression}

	Regression is a natural approach to estimate continuous variables, such as age, depth, and counts. 
	Some works
	suggest that regression is encouraged to be reformulated as an ordinal regression problem or a classification problem, which often enhances performance and benefits optimization~\cite{Cumulative_2013_CVPR,Ordinal_depth_2018_CVPR,li2018deep,Ordinal_age_2016_CVPR,liu2019classification}
	for many vision tasks. Ordinal regression is usually implemented by modifying well-studied classification algorithms and has been applied to the problem of age estimation~\cite{Ordinal_age_2016_CVPR} and monocular depth prediction~\cite{Ordinal_depth_2018_CVPR}. Li~\etal~\cite{li2018deep} further showed that directly reformulating regression to classification was also a good choice. In counting, the idea of blockwise classification is also investigated~\cite{liu2019classification}. All these attempts motivate us to devise a classification-based closed-set counter. In this work, in addition to the standard regression-based modeling as in~\cite{Lu2017TasselNet}, SS-DCNet also follows~\cite{li2018deep} and~\cite{liu2019classification} to discretize local counts and classify count intervals. Indeed, we observe in experiments that classification with S-DC
	generally works better than regression.
	
	\subsection{Open-Set Problems in Computer Vision}
	Many vision tasks are open-set by nature, such as depth prediction~\cite{Ordinal_depth_2018_CVPR,li2018deep}, age estimation~\cite{Cumulative_2013_CVPR,Ordinal_age_2016_CVPR}, object recognition~\cite{scheirer2012toward}, visual domain adaptation~\cite{panareda2017open}, etc. While the sense of the open set may be different, they generally suffer from poor generalization as object counting. However, we 
	find that, 
	the learning target of counting alone, i.e., the count value, can be easily transformed into a closed set (via spatial division).

	\section{Supervised Spatial Divide-and-Conquer Network}
	
	In this section, we describe how to construct a closed-set counter. We also explain our proposed SS-DCNet in detail.
	
	\subsection{Closed-Set Counter}\label{subsec:closed_set_counter}
	In local count modeling, there are two 
	approaches 
	to define a counter in the closed set $[0,C_{max}]$, i.e., counting by regression~\cite{Count_ception_2017_ICCVW,Lu2017TasselNet} and counting by classification~\cite{xhp2019SDCNet,liu2019classification}. In practice, $C_{max}$ should  not be greater than the maximum local count observed in the training set. It is clear that treating $C_{max}$ as the maximum prediction will cause a systematic error, but the error can be mitigated via S-DC, 
	as 
	discussed in Section~\ref{sec:exp}.
	
	\vspace{5pt}
	\noindent
	\textit{Regression-Based Counter (R-Counter):} R-Counter directly regresses count values within the closed set. If predicted count values are greater than $C_{max}$, the predictions will simply be truncated to $C_{max}$.
	
	\vspace{5pt}
	\noindent
	\textit{Classification-Based Counter (C-Counter):} Instead of regressing open-set count values, C-Counter discretizes local counts and classifies count intervals as in~\cite{liu2019classification}. Specifically, we define an interval partition of $[0,+\infty)$ as $\{0\}$, $(0,C_1]$, $(C_2,C_{3}]$, ... , $(C_{M-1},C_{max}]$ and $(C_{max},+\infty)$. These $M+1$ sub-intervals are labeled to the $0$-th to the $M$-th classes, respectively. For example, if a count value falls into $(C_2,C_{3}]$, it is labeled as the $2$-nd class. 
	The median of each sub-interval can be adopted when recovering the count from the interval. Notice that, for the last sub-interval $(C_{max},+\infty]$, $C_{max}$ will be used as the count value if a region is classified into this interval. 
	
	In what follows, we term the network 
	SS-DCNet (reg) when R-Counter is adopted, and SS-DCNet (cls) when C-Counter is used.
	
	\begin{figure*}[t]
		\begin{center}
			\includegraphics[width=.9\linewidth]{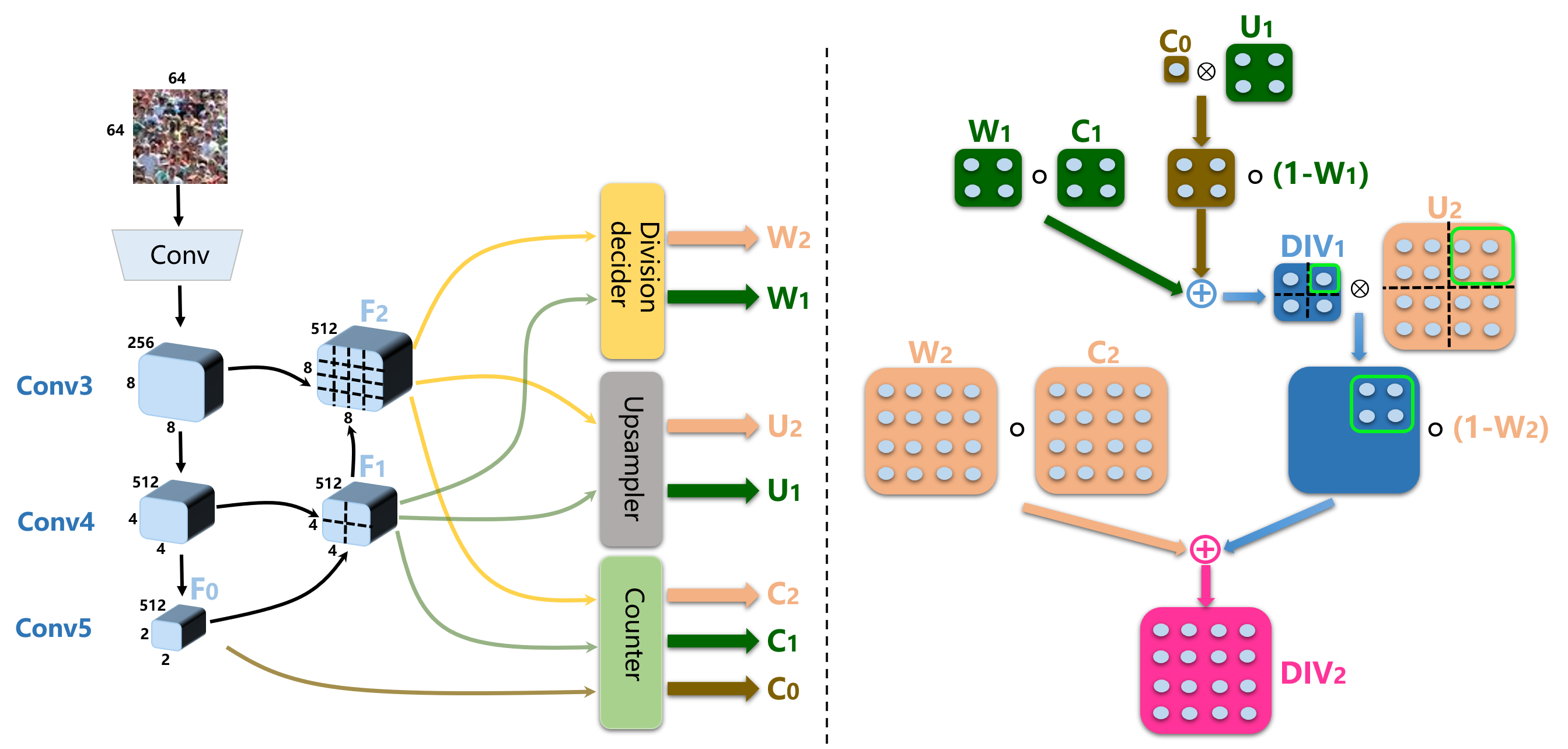}
		\end{center}
		\vspace{-10pt}
		\caption{ The architecture of SS-DCNet (left) and a two-stage S-DC process (right). SS-DCNet adopts all convolutional layers in VGG16~\cite{Simonyan2014Very_VGG16} (the first two convolutional blocks are simplified as $Conv$). An UNet~\cite{Unet2015U}-like decoder is employed to upsample and divide the feature map as 
		in 
		Fig.~\ref{fig:feature_divide}. A shared upsampler, a closed-set counter and a division decider receive divided feature maps, and respectively, generate upsampling map $U_i$s, division counts $C_i$s and division masks $W_i$s, for $i=1,2,...$. After obtaining these results, $C_{i-1}$ is upsampled with $U_i$, then it is merged with $C_i$ by $W_i$ to the $i$-th division count $DIV_i$ shown in the right sub-figure. In particular, we upsample each count of low resolution into the corresponding $2\times2$ area of high resolution before merging with $U_i$. ``$\circ$" denotes the Hadamard product, and ``$\otimes$'' denotes the Kronecker product. Note that, the $64\times64$ local patch is only used as an example for readers to understand the pipeline of SS-DCNet. Since SS-DCNet is a fully convolutional network, it can process an image of arbitrary sizes, say $M\times N$, and return $DIV_2$ of size $\frac{M}{64}\times \frac{N}{64}$. The configurations of the closed-set counter and the division decider are presented in Table~\ref{tab:classifer_div_decider}.}
		\label{fig:SDCNet_architecture}
	\end{figure*}

	\begin{table}\footnotesize
		\caption{%
		The Configurations of \textit{Counter}, \textit{Division decider} and \textit{Upsampler}. $AvgPool$ denotes Average Pooling. Convolutional layers are defined in the format:  $kernel~size$~Conv, $output~channel$, s~$stride$. Each convolutional layer is followed by ReLU except the last layer. In particular, a \textit{Sigmoid} function is attached at the end of \textit{division decider} to generate soft division masks. A \textit{Spatial Softmax} function is applied at the End of \textit{Upsampler}, which constrains the sum of upsampling weights in each $2\times 2$ adjacent regions to be $1$ and ensures consistent local count values in the same image area after upsampling.  The final output channel is $1$ for R-Counter and $class~num$ for C-Counter}
		\begin{center}
			\begin{tabular}{c|c}
				\hline
				Counter & Division decider/Upsampler\\
				\hline\hline
				
				$2\times2$ AvgPool, s $2$&  $2\times2$ AvgPool, s $2$ \\
				$1\times1$ Conv, $512$, s $1$ & $1\times1$ Conv, $512$, s $1$ \\
				$1\times1$ Conv, $1$/$\rm (class~num)$, s $1$ & $1\times1$ Conv, $1$, s $1$\\
				$-$&Sigmoid/Spatial Softmax\\
				
				\hline
			\end{tabular}
		\end{center}
		\label{tab:classifer_div_decider}
	\end{table}

	\subsection{Single-Stage Spatial Divide-and-Conquer}  \label{sec:S-DCNet_Arc}   
	As shown in
	Fig.~\ref{fig:SDCNet_architecture}, SS-DCNet includes a VGG16~\cite{Simonyan2014Very_VGG16} feature encoder, an UNet~\cite{Unet2015U}-like decoder, a closed-set counter, a division decider and an upsampler. The counter, the structures of division decider and the upsampler are shown in Table~\ref{tab:classifer_div_decider}. Note that, the first average pooling layer in the counter has a stride of $2$, so the final prediction has an output stride of $64$.

	The feature encoder removes fully-connected layers from the pre-trained VGG16. Suppose that the input patch is of size $64\times64$. Given the feature map $F_0$ (extracted from the Conv5 layer) with $\frac{1}{32}$ resolution of the input image, the counter predicts the local count value $C_0$ conditioned on $F_0$. Note that $C_0$ is the local count without S-DC, which is also the final output of previous approaches~\cite{Count_everyday_2017_CVPR,Count_ception_2017_ICCVW,Lu2017TasselNet}. %
	
	We execute the first-stage S-DC on the fused feature map $F_1$. $F_1$ is divided and sent to the shared counter to produce the division count $C_1\in\mathbb{R}^{2\times2}$. Concretely, $F_0$ is upsampled by $\times2$ in an UNet-like manner to $F_1$. Given $F_1$, the counter fetches the local features that correspond to spatially divided sub-regions, and predicts the first-level division counts $C_1$. Each of the $2\times2$ elements in $C_1$ denotes a sub-count of the corresponding $32\times32$ sub-region.
	
	With local counts $C_0$ and $C_1$, the next question is to decide where to divide. We learn such decisions with another network module, division decider, as 
	shown in the right part of Fig.~\ref{fig:SDCNet_architecture}. At the first stage of S-DC, the division decider generates a soft division mask $W_1$ of the same size as $C_1$ conditioned on $F_1$ such that for any $w\in W_1, w\in[0,1]$. $w=0$ means no division is required at this position, and the value in $C_0$ is used. $w=1$ implies that here the initial prediction should be replaced with the division count in $C_1$. Since both $W_1$ and $C_1$ are $2$ times larger than $C_0$, $C_0$ is required to be upsampled by $\times2$ to $\hat{C_0}$.
	
	Note  that, since $C_0$ denotes the local count of a $64\times64$ region, the sum of $\hat{C_0}$ should equal to $C_0$. The upsampling of $C_0$ is therefore a re-distribution operator that assigns $C_0$ to each sub-region. We compute the re-distribution map $U_1$ from the upsampler conditioned on $F_1$, and the sum of $U_1$ equals to $1$. We then upsample $C_0$ to $\hat{C_0}$ by  
	\begin{equation}\label{eq:upsample_c0}
	\hat{C_0} = (C_0 \otimes \bm{1}_{2\times 2}) \circ U_1\,,
	\end{equation}	
	where ``$\otimes$'' denotes Kronecker product and $\bm{1}_{2\times 2}$ denotes a $2\times 2$ matrix filled with $1$. Finally, the first-stage division result $DIV_1$ takes the form
	\begin{equation}\label{div1}
	DIV_1 = (\bm{1}-W_1)\circ \hat{C_0} + W_1 \circ C_1\,,
	\end{equation}
	where $\bm{1}$ denotes a matrix filled with $1$ and is with the same size of $W_1$, and ``$\circ$'' denotes the Hadamard product. 

	\begin{algorithm}[!t] \small
		\caption{ Multi-Stage S-DC}
		\label{alg:Overall algorith}
		\LinesNumbered
		\KwIn{Image $I$ and division time $N$}%
		\KwOut{Image count $C$}
		Extract $F_0$ from $I$\;
		Generate $C_0$ given $F_0$ with the closed-set counter\;%
		Initialize $DIV_0=C_0$\;
		\For{$i\leftarrow 1$ \textbf{to} $N$}
		{
			Decode $F_{i-1}$ to $F_i$\;
			Process $F_i$ with the closed-set counter, upsampler and the division decider to obtain $C_i$, $U_i$ and the division mask $W_i$\;
			Upsample $DIV_{i-1}$ to $\hat{DIV_{i-1}}$ with $U_i$ as
			Eq.~\eqref{eq:unsample12345}\;
			Update $DIV_i$ as 
			Eq.~\eqref{eq:merge_12345} \;  	
		}
		Integrate over $DIV_N$ to obtain the image count $C$\;
		\textbf{return} $C$
	\end{algorithm}

	\subsection{Multi-Stage Spatial Divide-and-Conquer}
	
	SS-DCNet can execute multi-stage S-DC by further decoding, dividing the feature map until reaching the output of the first convolutional block. In this sense, the maximum division time is $4$ in VGG16 for example. Actually we show later in experiments that a two-stage division is sufficient to 
	achieve 
	satisfactory performance. In multi-stage S-DC, $DIV_{i-1}$ ($i\ge2$) is first upsampled as:
	\begin{equation}\label{eq:unsample12345}
	\hat{DIV}_{i-1} = (DIV_{i-1}\otimes \bm{1}_{2\times 2}) \circ U_{i}\,,
	\end{equation}
	and then merged according to
	\begin{equation}\label{eq:merge_12345}
	DIV_i = (\bm{1}-W_i)\circ \hat{DIV}_{i-1} + W_i \circ C_i\,,
	\end{equation}
	in a recursive manner. Multi-stage SS-DCNet is summarized in Algorithm~\ref{alg:Overall algorith}.

	\subsection{Loss Functions}
	Here we elaborate the loss functions used in an $N$-stage SS-DCNet.

	\vspace{5pt}
	\noindent
	\textit{Counter Loss:}
	As mentioned in Section~\ref{subsec:closed_set_counter}, both R-Counter and C-Counter can be 
	used. 
	We 
	use the
	$\ell_1$ loss, denoted by $L_R^i$, $i=0,1,2,...N$, for each level of output $C_i$ when R-Counter is used, and cross-entropy loss, denoted by $L_C^i$, $i=0,1,2,...N$, when C-Counter is chosen. Note that, both ground-truth local counts and predicted counts are truncated to $C_{max}$ when R-Counter is adopted. The overall counter loss is $L_R=\sum_{i=0}^NL_R^i$ for the R-Counter and $L_C=\sum_{i=0}^NL_C^i$ for the C-Counter.

	\vspace{5pt}
	\noindent	
	\textit{Merging Loss (Implicit Division Supervision): } 
	We also adopt a $\ell_1$ loss $L_m$ for the final division output $DIV_N$. $L_m$ provides an implicit supervision signal for learning $W_i$s.
    
\begin{figure}[!t]
		\begin{center}
			\includegraphics[width=.9\linewidth]{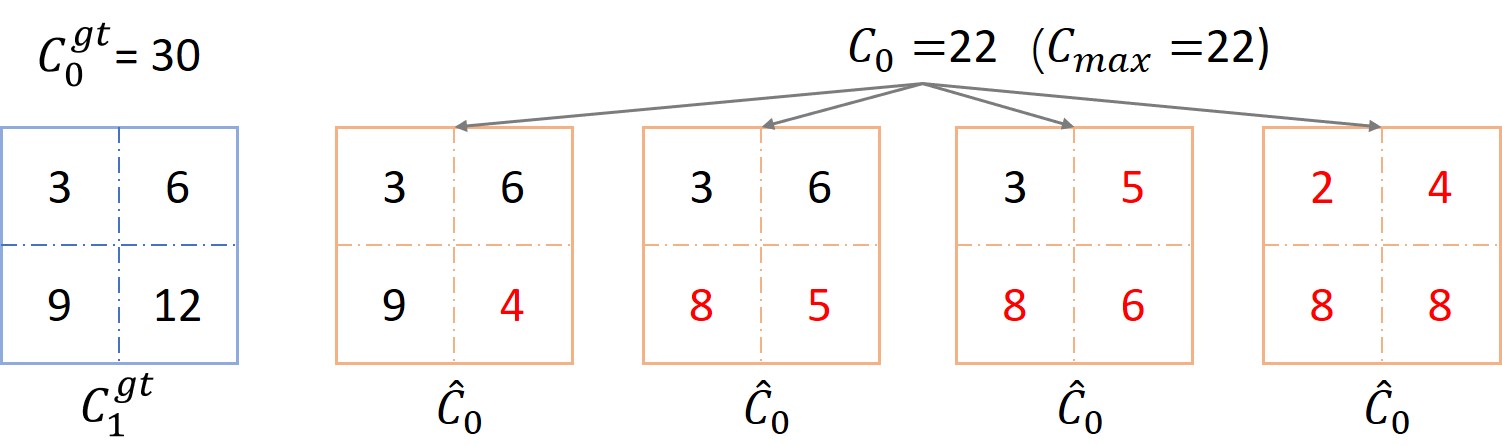}
		\end{center}
		\vspace{-10pt}
		\caption{Motivation of the division loss. When $C_0^{gt}>C_{max}$, the prediction $C_0$ can only be $C_{max}$ at most, and it is sure that at least one quarter of the region is underestimate.}
		\label{fig:Ldiv_motivation}
	\end{figure}

    \vspace{5pt}
	\noindent
	\textit{Division Loss (Explicit Division Supervision):} 
	We can also explicitly supervise $W_i$ by comparing the ground-truth $C_i^{gt}$ and $C_{max}$. As shown in Fig.~\ref{fig:Ldiv_motivation}, if the ground-truth count value of a $64\times 64$ local region $D$, i.e., $C_0^{gt}$, is larger than $C_{max}$, the inferred count $C_0$ will be no larger than $C_{max}$. Let us assume $C_0=C_{max}$. One knows that this local region is under-estimated ($C_{max}<C_0^{gt}$), but it does not imply that all sub-regions of $D$ are underestimated. %
	As shown in
	Fig.~\ref{fig:Ldiv_motivation}, there are $4$ possibilities where underestimations occur. In this case, we can only know that at least one sub-region of $D$ is underestimated and is required to be replaced with $C_1$, which means at least one of values of $W_1$ should approach $1$. Hence, we constrain this value to be the one with the largest probability approaching $1$ (the maximum) when $C_0^{gt}\textgreater C_{max}$.
	\begin{equation}\label{eq:div_loss_1}
	L_{div}^1 = - \mathbbm{1} \{C_0^{gt}>C_{max}\} \times \log(\max(W_1)) \,,
	\end{equation}  
	where $\mathbbm{1}\{P\}$ denotes the indicator function which outputs $1$ when the condition $P$ is true, and $0$ otherwise. $C_0^{gt}$ is the ground truth count value of $C_0$, and `$\max$' is the operator that returns the maximum value. 
	Following Eq.~\eqref{eq:div_loss_1}, the loss $L_{div}^i$, $i=1,2,...N$, for $W_i$ can be deduced as
	\begin{equation}\label{eq:div_loss_i}
	\begin{split}
	L_{div}^i &= -\sum_{j=1}^{H_i}\sum_{k=1}^{W_i} \mathbbm{1} \{C_{i-1}^{gt}[j,k]>C_{max}\} \\
	&\times \log(\max(W_{i}[2j-1:2j,2k-1:2k])) 
	\end{split}\,,
	\end{equation}
	where $C_{i-1}^{gt}[j,k]$ denotes the element of the $j$-th row and the $k$-th column of $C_{i-1}^gt$, and $W_{i}[2j-1:2j,2k-1:2k]$ the elements lying in the $(2j-1)$-th to $2j$-th row, $(2k-1)$-th to $2k$-th column of $W_{i}$.  
	The overall division loss $L_{div} = \sum_{i=1}^{N} L_{div}^i$.
	
	\begin{figure}[!t]
		\begin{center}
			\includegraphics[width=.9\linewidth]{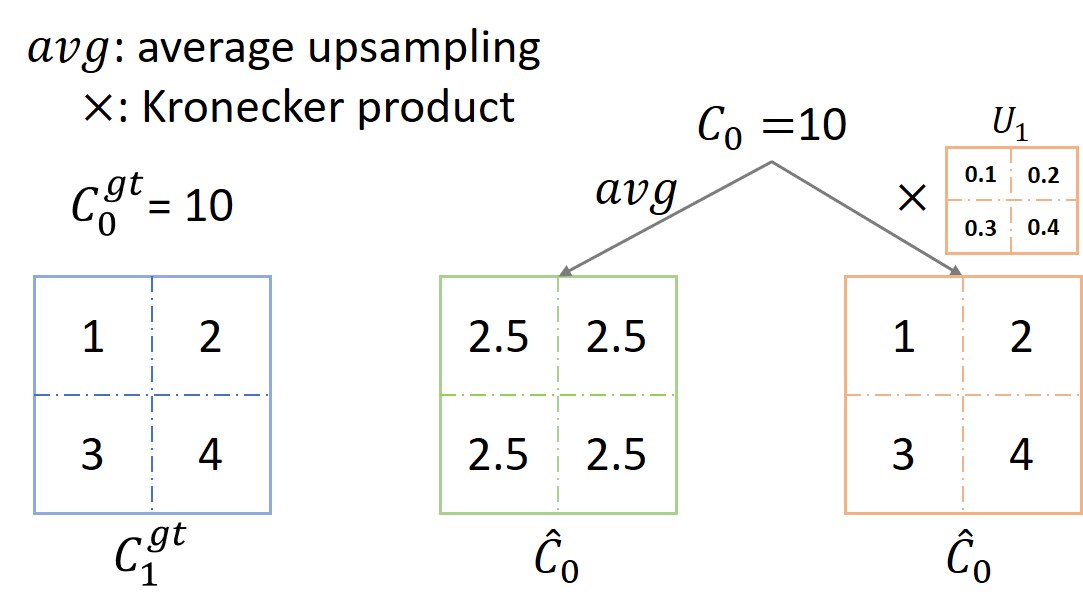}
		\end{center}
		\vspace{-10pt}
		\caption{
		Motivation of the upsampling loss. The middle depicts averaging upsampling %
		in S-DCNet~\cite{xhp2019SDCNet}, and the right shows guided upsampling in SS-DCNet informed by the upsampling map $U_1$. 
		Locally accurate $U_1$ leads to locally accurate $\hat{C}_0$.}
		\label{fig:Lup_motivation}
	\end{figure}
	
	\vspace{5pt}
	\noindent	
	\textit{Upsampling Loss:}
	By comparing count upsampling between S-DCNet~\cite{xhp2019SDCNet} and SS-DCNet in Fig.~\ref{fig:Lup_motivation}, we find that, even if $C_0$ is accurately predicted, average upsampling in S-DCNet can 
	produce 
	inaccurate $\hat{C}_0$. To alleviate such errors, we introduce an upsampling map $U_1$ to guide the upsampling of $C_0$. An upsampling loss is thus required to supervise the learning of $U_i$s. 
	
	The upsampler predicts $U_i$ used to re-distribute $DIV_{i-1}$ to its $\times2$ resolution output $\hat{DIV}_{i-1}$. We compute the ground truth of $U_i$ according to the distribution of local counts: 
	\begin{equation}\label{eq:u_i_GT}
	U_i^{gt} = C_i^{gt}/(C_{i-1}^{gt}\otimes \bm{1}_{2\times 2})\,,
	\end{equation} 
	where $U_i^{gt}$, $C_i^{gt}$ and $C_{i-1}^{gt}$ denote the ground truth of $U_i$, $C_i$ and $C_{i-1}$, respectively. `$/$' denotes element-wise division. We use $\ell_1$ loss $L_{up}^i$, $i=1,2,...N$, for each $U_i$. Hence, the overall upsampling loss $L_{up} = \sum_{i=1}^{N} L_{up}^i$.
	
	\vspace{5pt}
	\noindent
	\textit{Division Consistency Loss:} When R-Counter is used, we can further constrain the consistency between different $C_i$s when $C_i$s are in the range of the closed set $[0,C_{max}]$, which shares a similar spirit compared to~\cite{ACSCP_2018_CVPR}. For $C_0$ and $C_1$, the division consistency loss is defined by
	\begin{equation}\label{eq:L_same_1}
	L_{eq}^1   = \mathbbm{1}\{C_0^{gt}\leq C_{max}\}\times |C_0 - sum(C_1)|\,,
	\end{equation} 
	where `$sum$' is the operator that returns the sum of all elements.
	Following Eq.~\ref{eq:L_same_1}, the consistency loss between $C_{i-1}$ and $C_{i}$, $i=1,2,...N$, is
	\begin{equation}\label{eq:L_same_i}
	\begin{split}
	L_{eq}^i   &= \sum_{j=1}^{H_{i-1}} \sum_{k=1}^{W_{i-1}} \mathbbm{1}\{C_{i-1}^{gt}[j,k]\leq C_{max}\} \\
	&\times |C_{i=1}[j,k] - sum(C_i[2j-1:2j,2k-1:2k])| 
	\end{split}\,.
	\end{equation} 
	The overall division consistency loss $L_{eq}=\sum_{i=1}^{N} L_{eq}^i$.
	Note that, when C-Counter is adopted, the gradient of the consistency loss cannot be back propagated because count values are discretized into count intervals and represented by class labels. We simply drop the consistency loss in this case.

	As a summary, for SS-DCNet (reg) the final loss $L_{reg}$ is
	\begin{equation}
	L_{reg}=L_R+L_m+L_{up}+L_{div}+L_{eq}\,,
	\end{equation}  
	and for SS-DCNet (cls) the final loss $L_{cls}$ is
	\begin{equation}
	L_{cls}=L_C+L_m+L_{up}+L_{div}\,.
	\end{equation}

	\section{Open Set or Closed Set? A Theoretical Analysis}
	How does SS-DCNet benefit from transforming count values from open set to closed set? Here we first give the mathematical definitions of the open set, the closed set and the spatial division. With these definitions, we attempt to answer how many division times are required for transferring counts from the open set to the closed set in Proposition~\ref{prop:min_SDC_time}. Then in Proposition~\ref{prop:sdc_mae}, we show that, with sufficient spatial divisions, transforming count values from the open set can lead to lower absolute errors on the closed set, which sheds light on why we model counts in a closed set.

	\begin{definition}[Spatial Division of an Image]
		Given an image $I\in\mathbb{R}^{H\times W\times K}$, where $H$, $W$ and $K$ denote the height, width and channel dimensions, respectively, the spatial division of $I$ leads to a group of sub-images $\{I_i\in\mathbb{R}^{H_i\times W_i\times K}\}_{i=1,2,...,M}$ that satisfy:
		\\i) $I_i\subset I$;
		\\ii) $I_i\cap I_j = \phi$ , for $i\neq j$;
		\\iii) $I_1 \cup I_2 \cup ... \cup I_M = I$.
		\label{def:sdc_for_im}
	\end{definition}
	
	\begin{definition}[Open Set and Closed Set]
		Given a positive number $C_{max}$, for $\forall$ $x\geq0$ and $x\in\mathbb{R}$, we can define a closed set $\mathcal{S_C}$ by $\{x|0\leq x \leq C_{max}\}$ and an open set $\mathcal{S_O}$ by $\{x|x\textgreater C, C\geq C_{max} \}$. 
	\end{definition}
	
	In object counting,  $C_{max}$ is the maximum count value observed in the training set. Note that here we define the open set $\mathcal{S_O}$ to be $(C_{max},+\infty]$, rather than $[0,+\infty]$ aforementioned. This is because $[0,+\infty)$ and $\mathcal{S_C}$ only differ in the range of $(C_{max},+\infty]$, where S-DC is applied to this range to transform count values into $\mathcal{S_C}$. The shared interval $[0, C_{max}]$ remains unchanged and does not require S-DC. Hence, we define $S_O$ to be disjoint from $\mathcal{S_C}$ to simplify the analysis.     
	
	\begin{lemma}
		Given an image $I$, let $\boxplus$ be the spatial dichotomy division operator such that $\boxplus(I)=\{I_i\}_{i=1,2,3,4}$.
		Let $\boxplus^{N}$ further denote $\boxplus$ is applied for $N$ times. We have $\boxplus^{N}(I)= \{I_j\}_{j=1,2,...,4^N}$. 
		\label{lemma:max_division_parts}
	\end{lemma}	
	\begin{proof}
		Suppose that $M$ is the number of divided sub-images after $N$ divisions.
		\\\textit{i)} For $N=1$, according to the definition of $\boxplus$, 
		\begin{equation}
		\boxplus(I)=\{I_i\}_{i=1,2,3,4}\,,
		\end{equation}
		which means $M=4$;
		\\\textit{ii)} For $N=t$, assume $M=4^t$, we have 
		\begin{equation}
		\boxplus^t(I)=\{I_{k}\}_{k=4^t}\,,
		\end{equation} 
		then when $N=t+1$, for each $I_k\in \boxplus^t(I)$,
		\begin{equation}
		\begin{split}
		\boxplus^{(t+1)}(I) &= \boxplus(\boxplus^t(I))\\
		&=\{\boxplus(I_{k}),k=1,2,...,4^t\}\\
		&=\{I_{kp}\}_{k=1,2,...,4^t,p=1,2,3,4}
		\end{split}\,,
		\end{equation} 
		so $M=4\times 4^t=4^{t+1}$ holds for $N=t+1$. 
		
		Since both \textit{i)} and \textit{ii)} hold, by mathematical induction, we can deduce $M=4^N$ after $N$ divisions.
	\end{proof}
	
	According to Lemma~\ref{lemma:max_division_parts}, we know that, an image $I$ will be divided into $4^N$ sub-images at most after $N$ divisions. A subsequent question of interest is that, \textit{how many spatial divisions are required to transfer count values from $\mathcal{S_O}$ to $\mathcal{S_C}$? } This leads to our following proposition.
	
	\begin{proposition}[Minimum and Maximum Division Times]
		Assume an image $I\in\mathbb{R}^{H\times W\times K}$ with a count value $C^*\textgreater C_{max}$, $C^*\in\mathcal{S_O}$, is divided by the $\boxplus$ operator, and $r \times r$ is the minimum sub-region size with $C_{max}$ objects, then the required division times $N$ for transferring $C^*$ into $\mathcal{S_C}$ satisfy
		\begin{equation*}
		    \left \lceil \log_{4}{\frac{C^*}{C_{max} }} \right \rceil \leq N \leq \left \lfloor \max\{\log_{2}{\frac{H}{r}}, \log_{2}{\frac{W}{r}}\} \right \rfloor +1\,.
		\end{equation*}
	\label{prop:min_SDC_time}
	\end{proposition}
	\begin{proof}
		Suppose after $N$ division times, $I$ is divided into $M$ sub-images $\{I_i\in\mathbb{R}^{H_i\times W_i\times K}\}_{i=1,2,...,M}$ with local count values $c_i$s that satisfy \mbox{$0 \leq c_i\leq C_{max}$} and $\sum_{i=1}^M c_i=C^*$.
		
		i) Minimum Division Times. Since 
		\begin{equation}
		c_i\leq \max \limits_{i=1,2,..,M}c_i \leq C_{max} \,,
		\end{equation}
		we have
		\begin{equation}
		C^* = \sum_{i=1}^M c_i \leq M \times \max \limits_{i=1,2,..,M}c_i \leq  M \times C_{max} \,.
		\end{equation} 
		Hence, $M\geq \frac{C^*}{C_{max}}$. With Lemma~\ref{lemma:max_division_parts}, we know $M= 4^N$, so $4^N\geq \frac{C^*}{C_{max}}$, i.e., $N\geq \log_{4}{\frac{C^*}{C_{max} }}$.
		
		Note that, since $N$ is an integer, the minimum division times are $\left \lceil \log_{4}{\frac{C^*}{C_{max} }} \right \rceil$.
		
		ii) Maximum Division Times. First, we state that, if the size of all sub-images $I_i$ is less than $r\times r$ ($H_i<r$ and $W_i<r$ for $i=1,2,...,M$), then the counts $c_i$s of these sub-images will be less than $C_{max}$, i.e., $\max \limits_{i=1,2,...,M}c_i< C_{max}$. 
		
		We prove this statement with proof by contradiction. Suppose that there exists a sub-image $I_i$ of size $H_i\times W_i$ containing $c_i$ objects with $c_i\geq C_{max}$. Since $I_i$ has no less than $C_{max}$ objects, the minimum region size of $C_{max}$ objects 
		cannot exceed the size of $I_i$, i.e., $r\leq max\{H_i,W_i\}$.%
	     This contradicts with the assumption that $H_i<r$  and $W_i<r$. Hence, the assumption does not hold.

		Let $N_{r}$ denote the division times required to ensure that the sizes of all sub-images $I_i$s are no larger than $r\times r$, i.e., $N_r$ satisfies
	    \begin{equation}
	    \max\{\frac{H}{2^{N_{r}}},\frac{W}{2^{N_{r}}}\} < r\,.
	    \end{equation}
	    Since $N_{r}>\max\{\log_{2}{\frac{H}{r}}, \log_{2}{\frac{W}{r}}\}$ and $N_{r}$ is an integer, we have 
	    \begin{equation}
	        N_{r} = \left \lfloor \max\{\log_{2}{\frac{H}{r}}, \log_{2}{\frac{W}{r}}\} \right \rfloor +1\,.
	    \end{equation}
	    Hence, the required division times $N\leq N_{r}$.
	    
	    Proof completes.
	\end{proof}

	Proposition~\ref{prop:min_SDC_time} suggests the lower and higher bounds when SS-DCNet can transform count values from $\mathcal{S_O}$ to $\mathcal{S_C}$. This is a prerequisite that SS-DCNet can work. Proposition~\ref{prop:min_SDC_time} also allows one to have a prior estimate of the degree of granularity required in the spatial divisions. Is such a transformation effective? We further provide a theoretical analysis based on the metric of the absolute error. It is worth noting that the absolute error is widely considered to be an evaluation metric for counting, and we analyse the absolute counting error of the closed set and the open set with the help of the relative error. This is because that only the relative error normalized by the ground truth count can link the counting error across a wide range and provide a fair comparison between two distinct sets.    
	
	\begin{definition}
		Let $C$ ($C\textgreater 0$) be the ground-truth value, and $\hat{C}$ the inferred value. We define the relative error by $\epsilon_r=\frac{C-\hat{C}}{C}$ and the absolute error by $\epsilon_a = |C-\hat{C}| = C\times |\epsilon_r|$. \label{def:rel_err}
	\end{definition}
	
	By Definition~\ref{def:rel_err}, it is clear that the expectation of $|\epsilon_r|$ varies as the ground truth $C$ changes. We thus have
	\begin{definition}
		The function of the expectation w.r.t. $|\epsilon_r|$ is defined by $f(x)=\mathbb{E}_{C=x}|\epsilon_r|=\mathbb{E}_{C=x}|\frac{C-\hat{C}}{C}|$, and $f(x)$ is assumed to be continuous.
		\label{def:fx}
	\end{definition}

	Before
	presenting 
	our main results, we further need the conclusion of the following theorem.
	\begin{theorem}[Extreme Value Theorem~\cite{protter2012RealAnalysis}]
		If f(x) is a continuous function defined in the closed interval $[a,b]$, then $\exists$ $c\in [a,b]$ that satisfies $f(c)= \max \limits_{a \leq x \leq b}f(x)$. \label{the:closed_max}
	\end{theorem}

	According to Definition~\ref{def:rel_err}, Definition~\ref{def:fx} and Theorem~\ref{the:closed_max}, we arrive at our final proposition.
	\begin{proposition}\label{prop:sdc_mae}
		Let $\epsilon_a^o$ denote the absolute counting error on $\mathcal{S_O}$, $\epsilon_a^c$ the absolute counting error on $\mathcal{S_C}$ (after sufficient spatial divisions 
		as in 
		Proposition~\ref{prop:min_SDC_time}), $f(x)=\mathbb{E}_{C=x}|\epsilon_r|$, and $C_{max}$ a predefined positive number. Given a count value $C^*$, if $C^*>C_{max}$ and $ f(C^*) \textgreater \max \limits_{0 \leq x \leq C_{max}} f(x)$, then
		\begin{equation*}
		\mathbb{E}_{C=C^*}\left[\epsilon_a^c\right]\leq \max \limits_{0 \leq x \leq C_{max}} f(x)\times C^*<\mathbb{E}_{C=C^*}[\epsilon_a^o]\,.
		\end{equation*}
	\end{proposition}
	
	\begin{proof}
		For an image $I$ with a count value $C^*$, a spatial division of $I$, i.e., $\{I_i\}_{i=1,2,...,M}$, could be found in  Definition~\ref{def:sdc_for_im}. Their corresponding local counts $\{c_i\}_{i=1,2,...,M}$ satisfy \textit{i)} $0\leq c_i\leq C_{max}$, $i=1,2,...,M$, and \textit{ii)} $\sum_{i=1}^{M}c_i = C^*$.
		
		Let the $\epsilon_r^o$ denote the relative counting error on $\mathcal{S_O}$, and $\epsilon_r^i$ the relative counting error of each $I_i$ on $\mathcal{S_C}$. By Definition~\ref{def:rel_err}, we have
		\begin{equation}
		\epsilon_a^o = |C^* \times \epsilon_r^o| =  C^* \times |\epsilon_r^o|\,,
		\end{equation}
		and
		\begin{equation}
		\epsilon_a^c = \left|\sum_{i=1}^{M}c_i\times \epsilon_r^i\right|\,.
		\end{equation}
		By Definition~\ref{def:fx}, 
		\begin{equation}\label{eq:eao}
		\begin{split}
		\mathbb{E}_{C=C^*}[\epsilon_a^o] &= \mathbb{E}_{C=C^*}\left[C^* \times |\epsilon_r^o|\right]\\
		&=  C^* \times \mathbb{E}_{C=C^*}|\epsilon_r^o|\\
		&=C^* \times f(C^*)\
		\end{split}\,.
		\end{equation}
		\\With \textit{generalized triangle inequality}~\cite{triangle_ineq2011}, we have
		\begin{equation}\label{eq:mae-div}
		\epsilon_a^c=\left|\sum_{i=1}^{M}c_i\times \epsilon_r^i\right|\leq  \sum_{i=1}^{M}|c_i\times \epsilon_r^i| = \sum_{i=1}^{M}c_i\times |\epsilon_r^i|\,.
		\end{equation}
		By taking the expectation of both sides of Eq.~\eqref{eq:mae-div}, it amounts to
		\begin{equation}
		\begin{split}
		\mathbb{E}_{C=C^*}[\epsilon_a^c]&\leq \sum_{i=1}^{M}c_i\times \mathbb{E}_{C=C^*}|\epsilon_r^i|\\
		&= \sum_{i=1}^{M}c_i\times f(c_i)
		\end{split}\,.
		\end{equation}
		According to Theorem~\ref{the:closed_max}, $\exists~\zeta \in [0,C_{max}]$ such that \mbox{$f(\zeta)=\max \limits_{0 \leq x \leq C_{max}} f(x)$}. Hence, $f(c_i)\leq f(\zeta)$, for $i=1,2,...,M$, so
		\begin{equation}
		\begin{split}
		\mathbb{E}_{C=C^*}[\epsilon_a^c] &\leq \sum_{i=1}^{M}c_i\times f(c_i)\\ 
		&\leq   \sum_{i=1}^{M}c_i \times f(\zeta)  \\
		& =  f(\zeta) \times \sum_{i=1}^{M}c_i\\
		& = f(\zeta) \times C^*
		\end{split}\,.
		\end{equation}
		\\If $ f(C^*) \textgreater f(\zeta) = \max \limits_{0 \leq x \leq C_{max}} f(x)$, with Eq.~\eqref{eq:eao}, we have
		\begin{equation}
		\mathbb{E}_{C=C^*}[\epsilon_a^c] \leq f(\zeta) \times C^*\textless f(C^*)\times C^*= \mathbb{E}_{C=C^*}[\epsilon_a^o]\,.
		\end{equation}
		Proof completes.
	\end{proof}	

	Proposition~\ref{prop:sdc_mae} states that, by transforming the count value from $\mathcal{S_O}$ to $\mathcal{S_C}$, SS-DCNet can achieve lower counting errors on the condition that the expectation of the relative errors on $\mathcal{S_C}$ is smaller than that on $\mathcal{S_O}$. %
	We will verify this condition via experiments in Section~\ref{sec:toy}. According to Proposition~\ref{prop:sdc_mae}, it is encouraged to model counting in a closed set in theory.
	
	It is worth noting that, although our theory is developed specifically for object counting, the theoretical results are generic and not limited to this task. As long as the learning target is spatially divisible as the count value (so far we only find counting satisfies the property of spatial divisibility), without loss of generality, the same conclusion can be deduced, as stated in Corollary~\ref{col:sdcnet}.
	
	\begin{definition}[Spatial Divisibility]
		Let $P \in[0,+\infty)$ be the learning target of a vision task defined on an Image $I$. $\{I_i\}_{i=1,2,...,M}$ is the spatial division of $I$, and $p_i$ is the corresponding learning target of each $I_i$. $P$ is spatially divisible if $\sum_{i=1}^{M} p_i = P$.
		\label{def:spatially_divisible}  
	\end{definition} 
	
	\begin{corollary}
		Let $P$ be the learning target and is spatially divisible. Let $\epsilon_a^o$ denote the absolute error of $P$ on $\mathcal{S_O}$, $\epsilon_a^c$ the absolute error of $P$ on $\mathcal{S_C}$ (after spatial divisions), $f(x)=\mathbb{E}_{P=x}|\epsilon_r|$, and $P_{max}$ a predefined positive number.
		Given a positive number $P^*$, if $P^*>P_{max}$ and $ f(P^*) \textgreater \max \limits_{0 \leq x \leq P_{max}} f(x)$, then
		\begin{equation*}
		\mathbb{E}_{P=P^*}\left[\epsilon_a^c\right]\leq \max \limits_{0 \leq x \leq P_{max}} f(x)\times P^*<\mathbb{E}_{P=P^*}[\epsilon_a^o]\,.
		\end{equation*}
		\label{col:sdcnet}
	\end{corollary}

	\begin{figure*}[!t]
		\begin{center}
			\includegraphics[width=\linewidth]{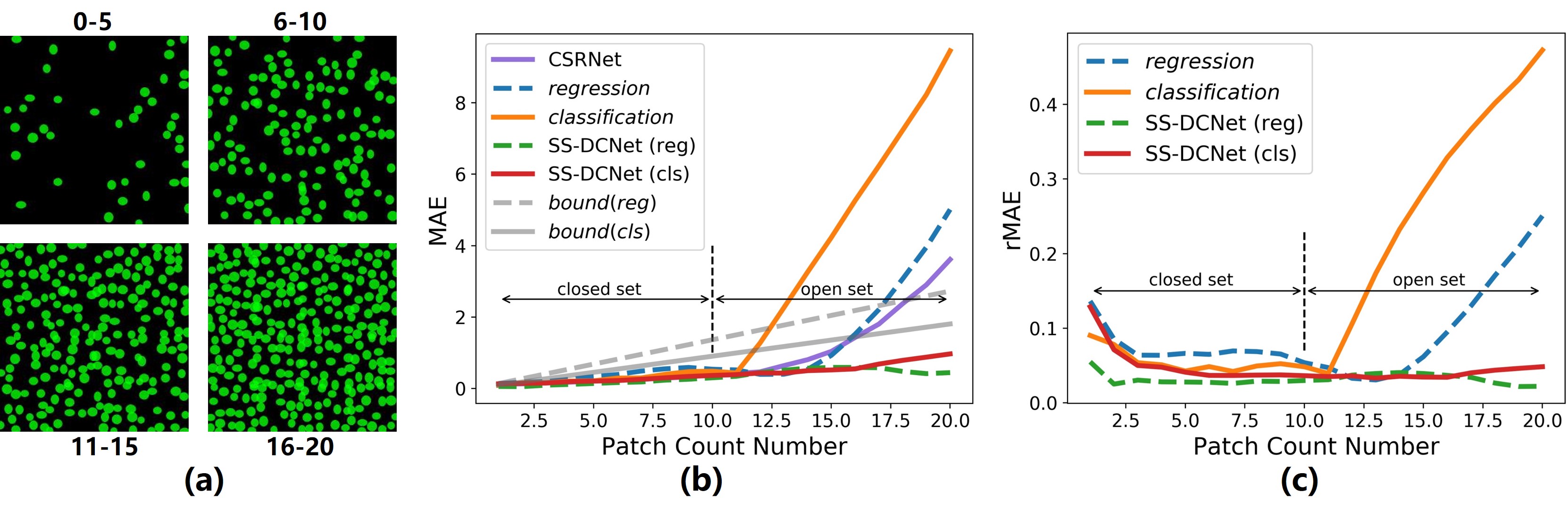}
		\end{center}
		\vspace{-20pt}
		\caption{A toy-level justification. (a) Some $256\times256$ images in the simulated cell counting dataset. The numbers denote the range of local counts of $64\times64$ sub-regions. (b) The mean absolute error (MAE) of different methods with increased $64\times64$ sub-region counts. (c) The relative mean absolute error (rMAE) of different methods with increased $64\times64$ sub-region counts. SS-DCNet (reg/cls) adopts one-stage S-DC.
		}
		\label{fig:simulated_d_t}
	\end{figure*}
	
	\section{Open Set or Closed Set? A 
	Justification
	on a Synthetic dataset
	}\label{sec:toy}
	As aforementioned, counting is an open-set problem, while the model is learned in a closed set. \emph{Can a closed-set counting model really generalize to open-set scenarios?} Here we show through a controlled toy experiment that, the answer is \textit{negative}. In addition, in this experiment we illustrate that SS-DCNet indeed works better than that without S-DC, which supports our Proposition~\ref{prop:sdc_mae}. %
	Inspired by~\cite{vlaz2010denlearn}, we synthesize a cell counting dataset to explore the counting performance outside a closed training set.
	
	\subsection{
	Synthetic 
	Cell Counting Dataset}
	We first generate $500$ $256\times256$ images with $64\times64$ sub-regions containing only $0\sim10$ cells to construct the training set (a closed set). To generate an open testing set, we further synthesize $500$ images with sub-region counts uniformly distributed in the range of $[0,20]$. 

	\subsection{Baselines and Protocols}
	We 
    implement 
	three approaches for comparisons, %
	which 
	are:
	\textit{i}) a density regression baseline CSRNet~\cite{CSRNet_2018_CVPR};	
	\textit{ii}) a regression baseline with pretrained VGG16 as the backbone and the R-Counter used in SS-DCNet as the backend, without S-DC. $\ell_1$ loss is used. This baseline directly regresses the open-set counts;
	\textit{iii}) a classification baseline with the same VGG16 and the C-Counter, without S-DC; %
	\textit{iv}) our proposed SS-DCNet, which learns from a closed set but adapts to the open set via S-DC. According to Proposition~\ref{prop:min_SDC_time}, at least $1$-time division is required for SS-DCNet to transform count values from the open set to the closed set. We adopt both SS-DCNet (reg) and SS-DCNet (cls) with $1$-time division for comparison.

	As for
	the discretization of count intervals, we choose $0.5$ as the step because cells may be overlapping in local patches. Hence, we have a partition of $\{0\}$, $(0.0.5]$,$(0.5,1]$, ... ,$(9.5,10]$ and $(10,+\infty)$. All approaches are trained with standard stochastic gradient descent (SGD). The learning rate is initially set to $0.0001$ and is decreased by $\times 10$ when the training error stagnates.
	
	\subsection{Observations}
	According to Fig.~\ref{fig:simulated_d_t}($b$), it can be observed that both regression and classification baselines work well in the range of the closed set ($0\sim10$), but the counting error increases 
	quickly
	when counts are larger than $10$. This suggests
	that 
	a conventional counting model learned in a closed set cannot generalize to the open set. However, SS-DCNet can achieve accurate predictions even on the open set, which confirms the advantage of S-DC.

	\subsection{Analyses}
	The relative mean absolute error (rMAE) is an empirical estimate of $f(x)$ 
	according to 
	Definition~\ref{def:fx}, and the MAE is also an empirical estimate of $\mathbb{E}_{C=C_0}\{\epsilon_a\}$. Fig.~\ref{fig:simulated_d_t}($c$) and ($b$) report how these two metrics vary, respectively. We have the following discussions:
	\squishlist
	\item As shown in Fig.~\ref{fig:simulated_d_t}($c$), $ f(C_0) \textgreater \max \limits_{0 \leq x \leq C_{max}} f(x)$ satisfies for the classification baseline when the patch count $C_0\geq 12$. According to Proposition~\ref{prop:sdc_mae}, under the condition above, SS-DCNet (cls) will show lower $MAE$ than the classification baseline without S-DC. When $C_0\geq17$, the same conclusion can be drawn between SS-DCNet (reg) and its open-set regression baseline.
	\item When $C_0\in[10,12]$, $ f(C_0) \textgreater \max \limits_{0 \leq x \leq C_{max}} f(x)$ is no longer true for the classification baseline. As shown in Fig.~\ref{fig:simulated_d_t}($b$), SS-DCNet (cls) only reports comparable results against the classification baseline.  When $C_0\in[10,17]$, the same observation can be made between SS-DCNet (reg) and its open-set regression counterpart.
	\squishend
	
	In general, our %
	experiment on the synthetic data verifies Proposition~\ref{prop:sdc_mae} to 
	some 
	extent. According to these results, it is also encouraged to model counting in a closed set in practice. 

	\section{Experiments on Real%
	Datasets}\label{sec:exp}

	Extensive experiments are further conducted to demonstrate the effectiveness of SS-DCNet on real%
	datasets. We first describe some essential implementation details. %
	Then ablation studies are conducted on the ShanghaiTech Part\_A~\cite{MCNN_2016_CVPR} dataset to highlight the benefit of S-DC. We then compare SS-DCNet against current state-of-the-art methods on five public datasets. Finally, we also report cross-domain performance to verify the generalization ability of SS-DCNet.

	Mean Absolute Error ($MAE$) and Root Mean Squared Error ($MSE$) are chosen to quantify the counting performance. They are defined by
	\begin{equation}\label{MAE}
	\small
	MAE=\frac{1}{Z} \sum_{i=1}^{Z} |C^{pre}_{i}-C^{gt}_{i}|\,,
	\end{equation}
	\begin{equation}\label{RMSE}
	\small
	MSE=\sqrt{ \frac{1}{Z} \sum_{i=1}^{Z} (C^{pre}_{i}-C^{gt}_{i})^{2}}\,,
	\end{equation}
	where $Z$ denotes the number of images, $C^{pre}_{i}$ denotes the predicted count of the $i$-th image, and $C^{gt}_{i}$ denotes the corresponding ground-truth count. $MAE$ measures the accuracy of counting, and $MSE$ measures the stability. Lower $MAE$ and $MSE$ imply better counting performance. 
	
	In addition, the absolute error is not always
	meaningful, 
	because a mistake of $1$ for a ground truth count of $2$ might seem egregious but the same mistake for the ground truth count of $23$ might seem reasonable. This is rooted in the fact that human perception of count is essentially logarithmic and not linear~\cite{dehaene2008log}. Aside from $MAE$ and $MSE$, we further report the relative Mean Absolute Error (rMAE) for most datasets used, defined by
	\begin{equation}\label{rMAE}
	\small
	rMAE=\frac{1}{Z} \sum_{i=1}^{Z} \frac{|C^{pre}_{i}-C^{gt}_{i}|}{C^{gt}_{i}}\,,
	\end{equation}

	\subsection{Implementation Details}
	
	\subsubsection{Interval Partition for C-Counter}\label{One_two_linear}
	
	We generate ground-truth counts of local patches by integrating over the density maps. The counts are usually not integers, because objects can partly present in cropped local patches.  
	We evaluate two different partition strategies. In the first partition, we choose $0.5$ as the step and generate partitions as $\{0\}$, $(0,0.5]$, $(0.5,1]$, ..., $(C_{max}-0.5,C_{max}]$ and $(C_{max},+\infty)$, where $C_{max}$ denotes the maximum count of the closed set. This partition is named as \texttt{One-Linear Partition}.
	
	In the second partition, we further finely divide the sub-interval $(0,0.5]$, because this interval contains a sudden change from no object to part of an object, and a large proportion of objects lie in this sub-interval. A small step of $0.05$ is further used to divide the sub-interval $(0,0.5]$, i.e., $(0,0.05]$, $(0.05,0.1]$, ..., $(0.45,0.5]$. Other intervals remain the same as \texttt{One-Linear Partition}. We call this partition \texttt{Two-Linear Partition}.

	\subsubsection{Data Preprocessing}\label{subsec:data_pre_process}

	We follow the same data augmentation used in~\cite{CSRNet_2018_CVPR}, except for the UCF-QNRF dataset~\cite{Compose_Loss_2018_ECCV} where we adopt two data augmentation strategies. In particular, $9$ sub-images of $\frac{1}{4}$ resolution are cropped from the original image. The first $4$ sub-images are from four corners, and the remaining $5$ are randomly cropped. Random scaling and flipping are also executed. %
	
	\subsubsection{Training Details}
	SS-DCNet is implemented with \texttt{PyTorch}~\cite{paszke2019pytorch}. We train SS-DCNet using SGD. The encoder in SS-DCNet is directly adopted from convolutional layers of VGG16~\cite{Simonyan2014Very_VGG16} pretrained on ImageNet, and the other layers employ random Gaussian initialization with a standard deviation of $0.01$. The learning rate is initially set to $0.001$ and is decreased by $\times10$ when the training error stagnates. We keep training until convergence. For the ShanghaiTech, UCF\_CC\_50, TRANCOS and MTC datasets, the batch size is set to $1$. For the UCF-QNRF dataset, the batch size is set to 16 following~\cite{Compose_Loss_2018_ECCV}. 
	
	\subsection{Ablation Study on the ShanghaiTech Part\_A}

	\subsubsection{Is SS-DCNet Robust to $C_{max}$?}
	When reformulating the counting problem into classification, a critical issue is how to choose $C_{max}$, which defines the closed set. Hence, it is important that SS-DCNet is robust to the choice of $C_{max}$.
	
	We conduct a statistical analysis on count values of local patches in the training set, and then set $C_{max}$ with the quantiles ranging from $100\%$ to $80\%$ (decreased by $5\%$). Two-stage SS-DCNet is evaluated. Another baseline of classification without S-DC is also used to explore whether counting can be simply modeled in a closed-set classification manner. To be specific, we reserve the VGG16 encoder and the C-Counter in this classification baseline.
	
\begin{figure}[t]
		\begin{center}
			\includegraphics[width=1.0\linewidth]{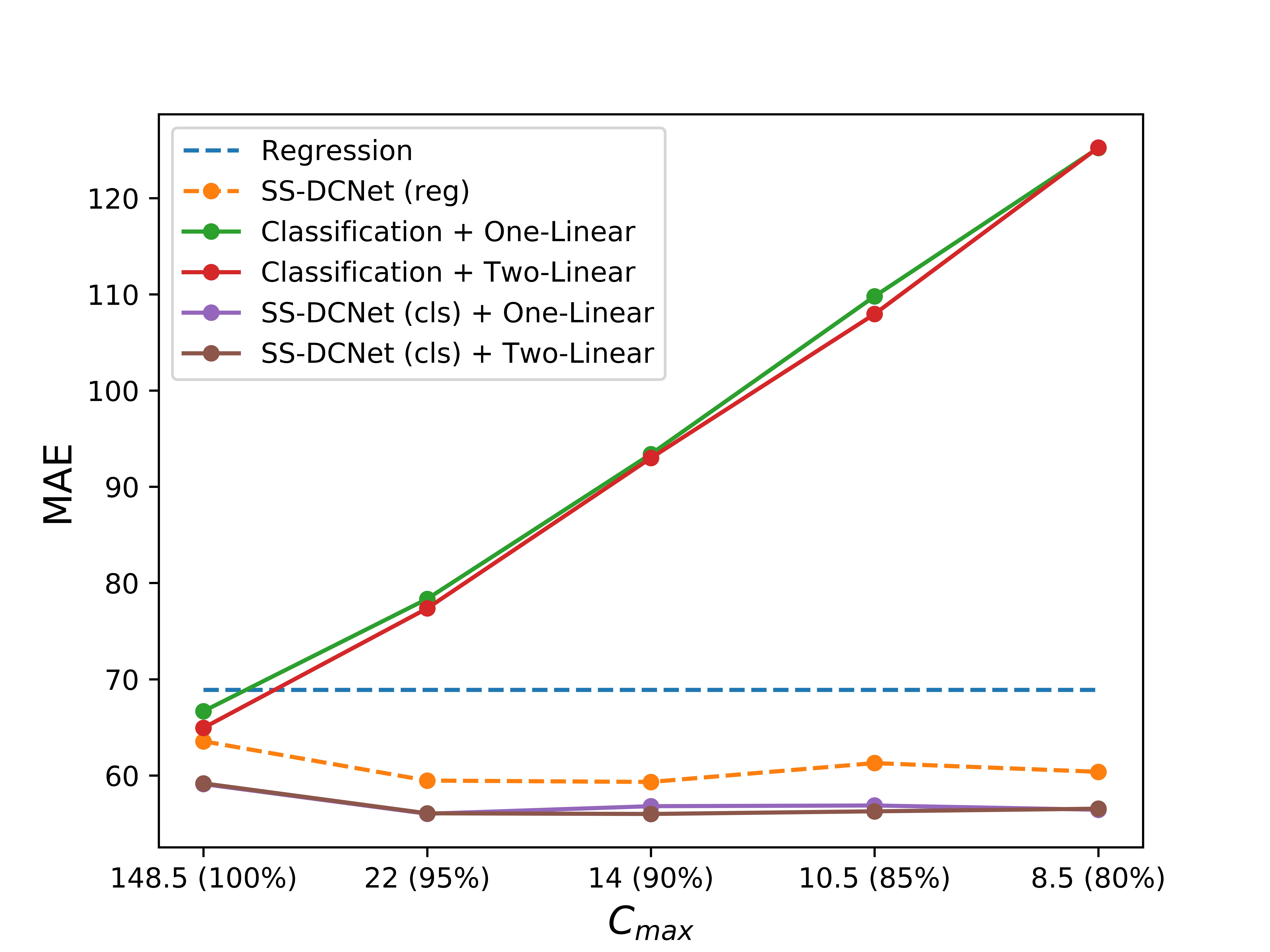}
		\end{center}
		\vspace{-10pt}
		\caption{The influence of $C_{max}$ to SS-DCNet on the ShanghaiTech Part\_A dataset~\cite{MCNN_2016_CVPR}. 
			The numbers in the brackets denote quantiles of the training set, for example, $22$ ($95\%$) means the $95\%$ quantile is $22$. `VGG16~Encoder' is the classification baseline without \mbox{S-DC}. 
			`One-Linear' and `Two-Linear' are defined in Section~\ref{One_two_linear}. SS-DCNet (reg/cls) adopts two-stage S-DC.}
		\label{fig:mae_versus_cmax}
	\end{figure}

	Results are presented in Fig.~\ref{fig:mae_versus_cmax}. 
	We see
	that the MAE of the classification baseline increases rapidly with decreased $C_{max}$. This result is not surprising, because the model is constrained to be visible to count values not greater than $C_{max}$.  This suggests that counting cannot be simply transformed into closed-set classification. However, with the help of S-DC, SS-DCNet exhibits strong robustness to the changes of $C_{max}$. It seems that the systematic error brought by $C_{max}$ can somewhat be alleviated with S-DC. 
	As for
	how to choose %
	a proper value for 
	$C_{max}$, the maximum count of the training set seems not the best choice, while setting to  some smaller
	values
	even delivers better performance. 
	It may be due to that 
	a model is only able to count objects accurately within a certain degree of denseness. %
	We also notice that
	\texttt{Two-Linear Partition} is slightly better than \texttt{One-Linear Partition}, which indicates that the fine division to the $(0,0.5]$ sub-interval has a positive effect.

	\begin{figure}
		\makeatletter\def\@captype{table}\makeatother
			\caption{%
			Results of SS-DCNet with different S-DC stages. 
			The best performance is in boldface.}
			\centering
			\footnotesize
			\addtolength{\tabcolsep}{-1pt}
			\begin{tabular}{c|ccc|ccc}
				\hline
				&\multicolumn{3}{c|}{SS-DCNet (cls)} &\multicolumn{3}{c}{SS-DCNet (reg)} \\
				\cline{2-3} \cline{4-5}\cline{6-7}
				Division time& MAE &MSE&rMAE & MAE &MSE&rMAE\\
				\hline
				0 & 76.0 & 142.5&16.48\%&76.7&144.6&16.41\%\\		  
				1 & 57.8 & 92.0&13.81\% &61.0&98.1&14.68\%\\
				2 & \textbf{56.1} & \textbf{88.9}&\textbf{13.78\%} &\textbf{59.5}  &\textbf{95.0}&\textbf{14.35}\%\\
				3 & 57.0 & 92.7&13.98\% &60.1&97.2&14.37\%\\	 
				4 & 59.1 & 100.0&14.23\% &62.8&99.1&15.39\%\\
				\hline
			\end{tabular}
			\label{tab:div_time}
	\end{figure}
	
	\begin{figure*}
		\begin{minipage}{\textwidth}
			\begin{minipage}{0.3\textwidth}
				\makeatletter\def\@captype{table}\makeatother
				\caption{
				Effect of S-DC. The best performance is in  boldface.}
				\centering
				\footnotesize
				\begin{tabular}{l|ccc}
					\hline
					Method           & MAE &MSE&rMAE\\
					\hline
					
					classification   & 77.4 &  149.3&17.13\%  \\
					$S_c$~regression	 &76.5 &  140.9&16.80\%\\   
					$S_o$~regression  & 68.9 & 112.1& 16.43\%  \\
					SS-DCNet (reg)   &59.5  &95.0&14.35\% \\
					SS-DCNet (cls) & \textbf{56.1}& \textbf{88.9}&\textbf{13.78\%}\\
					
					\hline
				\end{tabular}
				\label{tab:effect of S-DC} 
			\end{minipage}
			\hfill
			\makeatletter\def\@captype{table}\makeatother
			\begin{minipage}{0.65\textwidth}
				\caption{
				Effect of different loss functions. }
				\vspace{-10pt}
				\centering
				\footnotesize
				\begin{tabular}{l|cccccc|cccccccc}
					\hline

					Method  &$L_R$&$L_C$&$L_m$&$L_{up}$& $L_{div}$&$L_{eq}$ &MAE &MSE&rMAE\\
					
					\hline
					S-DCNet (reg)~\cite{xhp2019SDCNet} &\checkmark&&\checkmark&&&         &64.7  & 105.7&15.84\%\\
					\hline
					\multirow{3}{*}{SS-DCNet (reg)}&\checkmark&&\checkmark&\checkmark&       &&61.5	 & 	99.2&15.34\%	\\
					&\checkmark&&\checkmark&\checkmark&\checkmark&         &60.6   &96.5&15.46\% \\
					&\checkmark&&\checkmark&\checkmark&\checkmark&\checkmark       &59.5  &95.0&14.35\% \\
					\hline
					S-DCNet (cls)~\cite{xhp2019SDCNet}  &&\checkmark&\checkmark&&      && 58.3 & 95.0&13.94\%\\
					\hline
					
					\multirow{2}{*}{SS-DCNet (cls)}   &&\checkmark&\checkmark&\checkmark&      &&57.8 &100.8&13.92\%  \\
					&&\checkmark&\checkmark&\checkmark&\checkmark      &&\textbf{56.1}&\textbf{88.9}&\textbf{13.78\%} \\
					\hline
				\end{tabular}
				\label{tab:reg_cls_div} 
			\end{minipage}
			
		\end{minipage}
	\end{figure*}	
	
	According to the results above, SS-DCNet is robust to $C_{max}$ in a wide range of values, and $C_{max}$ is generally encouraged to be set less than the maximum count value observed. In addition, there is no significant difference between two kinds of partitions. For simplicity, we set $C_{max}$ to be the $95\%$ quantile and adopt \texttt{Two-Linear Partition} in the following experiments.

	\subsubsection{How Many Times to Divide?}
	SS-DCNet can apply S-DC by up to $4$ times, but how many times are sufficient? Here we evaluate SS-DCNet with different division stages. The maximum count value of $64\times64$ image patches in the test set is $136.50$ and $C_{max}=22$. With Proposition~\ref{prop:min_SDC_time}, we know 
	twice 
	division is required at least. Quantitative results are listed in Table~\ref{tab:div_time}. It can be observed that when the division time $N$ varies from $0$ to $2$, the counting error $MAE$ and $rMAE$ significantly decreases for both SS-DCNet (reg) and SS-DCNet (cls). However, counting accuracy saturates when $N$ continues increasing. In general, two-stage S-DC seems sufficient. We use this setup in the following experiments.
	
	\subsubsection{The Effect of S-DC}	
	To highlight the effect of S-DC, we compare SS-DCNet against several regression and classification baselines. These baselines adopt the same architecture of VGG16 encoder and the counter in SS-DCNet. \textit{classification} is the result of $C_0$ adopting C-Counter without S-DC, and $C_{max}$ is set to be the $95\%$ quantile ($C_{max}=22$). For regression baselines, we employ R-counter to obtain the prediction $C_0$ without S-DC.  We create two regression baselines. \textit{open-set regression + S-DC} is straightforward. We do not limit the output range, and it can vary from $0$ to $+\infty$.  \textit{$S_c$ regression} indicates that the output range is constrained within $[0,C_{max}]$  ($C_{max}$ is also set to $22$ for a fair comparison). Any large outputs will be clipped to $C_{max}$.
	
	Results are shown in Table~\ref{tab:effect of S-DC}. We can see that counting by classification without S-DC suffers from the limitation of $C_{max}$ and performs even worse than $S_o$ regression. $S_c$ regression also suffers from the same problem. However, with S-DC, SS-DCNet (reg/cls) significantly reduces the counting error and outperforms both their regression/classification baseline by a large margin. It suggests that a counting model can learn from a closed set and generalize well to an open set via S-DC. We notice that SS-DCNet (cls) performs better than SS-DCNet (reg). It seems that reformulating counting in classification is more effective than in regression. One plausible reason is that the optimization is easier and less sensitive to sample imbalance in classification than in regression.

\begin{figure}[!t]
		\begin{center}
			\includegraphics[width=.9\linewidth]{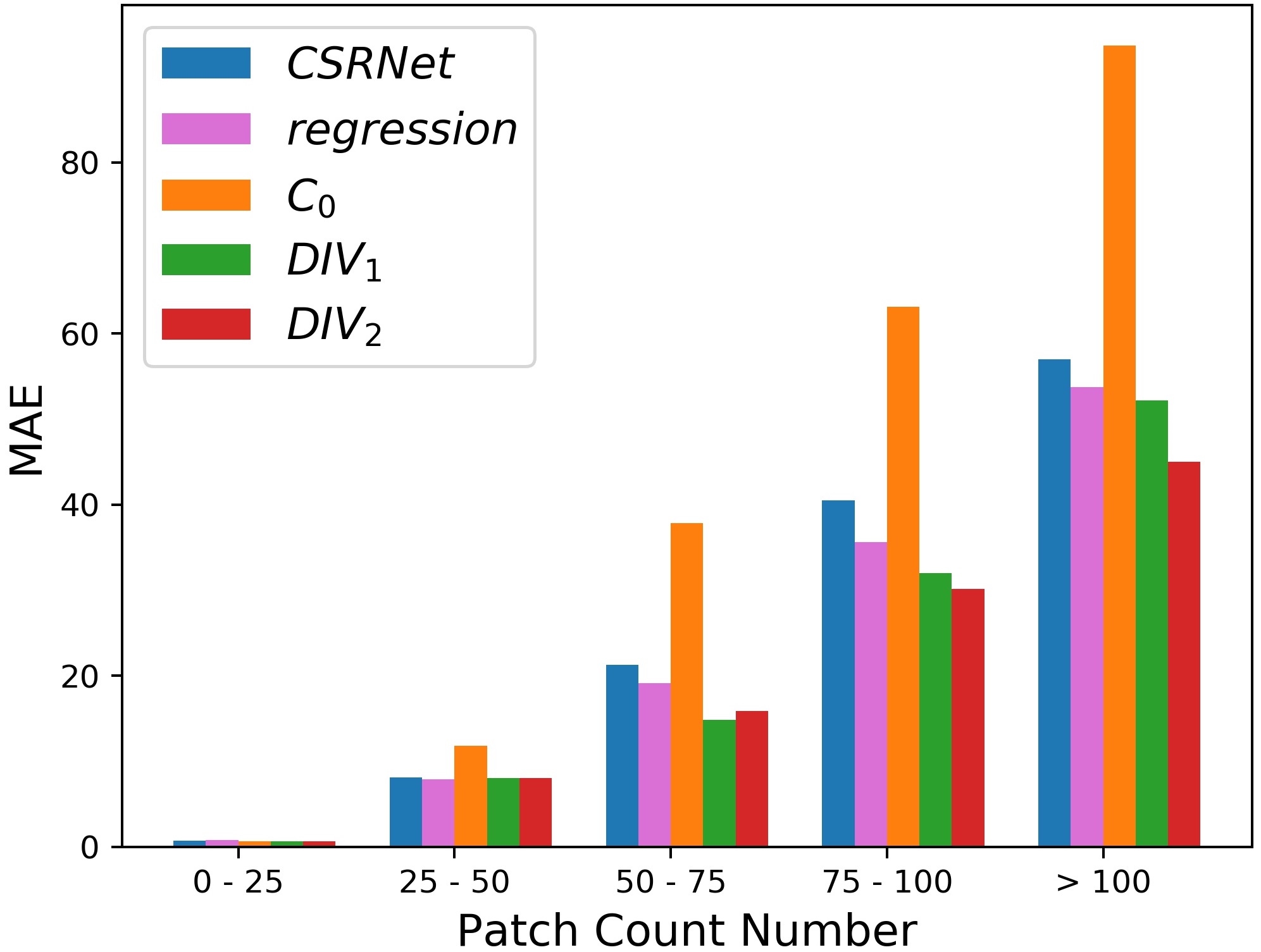}
		\end{center}
		\vspace{-20pt}
		\caption{Counting errors of $64\times64$ local patches on the test set of ShanghaiTech Part\_A~\cite{MCNN_2016_CVPR}. $CSRNet$~\cite{CSRNet_2018_CVPR} is a density map regression method which adopts VGG16~\cite{Simonyan2014Very_VGG16} as the feature extractor. $regression$ denotes direct open-set local counts regression using VGG16 ($S_o$ regression). $C_0$, $DIV_1$ and $DIV_2$ are count value predictions conditioned on $F_0$, $1$-stage division and $2$-stage division of SS-DCNet (cls), respectively. }
		\label{fig:patch_div_err}
	\end{figure}

	We further analyze the counting error of $64\times64$ local patches in detail. As shown in Fig.~\ref{fig:patch_div_err}, we observe that the direct prediction $C_0$ without S-DC %
	performs worse than the $S_o$ regression baseline and CSRNet, which can be attributed to the limited $C_{max}$ of the C-Counter. After embedding S-DC, the counting errors ($MAE$ and $rMAE$) of $DIV_1$ and $DIV_2$ significantly reduce and outperform open-set regression and CSRNet. Such a benefit is even much clear in dense patches with local counts greater than $100$. It justifies our argument that, instead of regressing a large count value directly, it is more accurate to count dense patches through S-DC, which verifies the conclusion in Proposition~\ref{prop:sdc_mae} in the real-world dataset.

	\subsubsection{Choices of Loss Functions}
	Here we validate the effect of different loss functions used in SS-DCNet. Results are reported in Table~\ref{tab:reg_cls_div}. As analyzed in S-DCNet~\cite{xhp2019SDCNet}, $L_C$ provides supervision to $C_i$s, and $L_m$ implicitly supervises the division weights $W_i$s. With only $L_C$ and $L_m$, S-DCNet (cls) can achieve good division results. However, S-DCNet (reg) cannot report competitive results as S-DCNet (cls) with $L_R$ and $L_m$. After incorporating the upsampling loss $L_{up}$ in the SS-DCNet (reg/cls), MAE reduces by $3.2$/$0.5$ and rMAE reduces by $0.5\%$/$0.02\%$. Such an improvement can be attributed to the replacement of the average upsampling in S-DCNet with learned upsampling in SS-DCNet. $L_{div}$ provides explicit supervision for spatial division weight $W_i$s. One can see that, $L_{div}$ can further improve the counting performance of SS-DCNet (reg/cls), and clear division results can be observed as shown in Fig.~\ref{fig:w_visual}. Moreover, division consistency loss $L_{eq}$ is also effective for SS-DCNet (reg), with $1.1$ improvement in MAE and $1.11\%$ in improvement in rMAE. Overall, SS-DCNet (reg/cls) shows a clear advantage over its previous version S-DCNet~\cite{xhp2019SDCNet} with the help of additional supervision $L_{up}$, $L_{div}$ and $L_{eq}$. 
	
	\begin{table}[!t]\footnotesize
		\caption{
		Configuration of SS-DCNet. $max$ denotes the maximum count of local patches
		in the training set. $C_{max}$ is the maximum count of the closed set in SS-DCNet. $Gaussian~kernel$ is used to generate density maps from dotted annotations. Specially, since UCF\_CC\_50 adopts 5-fold cross-validation, $max$ and $C_{max}$ are set adaptively for each fold.}
		\centering
		\addtolength{\tabcolsep}{4pt}
		\begin{tabular}{l|c|c|cc}
			\hline
			Dataset & $C_{max}$ &max& Gaussian kernel\\
			\hline
			SH Part\_A~\cite{MCNN_2016_CVPR}&22.0	& 148.5 &\multirow{3}{*}{Geometry-Adaptive}\\ \cline{1-3} 
			UCF\_CC\_50~\cite{UCFCC50_2013_CVPR}& $-$ & $-$ &\\	\cline{1-3} 
			UCF-QNRF~\cite{Compose_Loss_2018_ECCV}& 8.0 & 131.5& \\ \cline{1-4}
			SH Part\_B~\cite{MCNN_2016_CVPR}& 7.0 & 83.0&  Fixed: $\sigma=15$\\ \cline{1-4}
			Trancos~\cite{TRANCOSdataset_IbPRIA2015}& 5.0 & 24.5& Fixed: $\sigma=10$\\	   \cline{1-4}
			MTC~\cite{Lu2017TasselNet}& 3.5 & 8.0& Fixed: $\sigma=8$\\	   \cline{1-4}
			\hline
			Partition& \multicolumn{3}{c}{Two-Linear}\\ \cline{1-4}
			Type of~$C_{max}$ &\multicolumn{3}{c}{$95\%$ quantile}\\ \cline{1-4}
			\hline
		\end{tabular}
		\label{tab:compare_setting}
	\end{table}

	\subsubsection{Spatial Divide-and-Conquer versus Spatial Attention}
\begin{figure*}[!t]
		\begin{center}
			\includegraphics[width=1.00\linewidth]{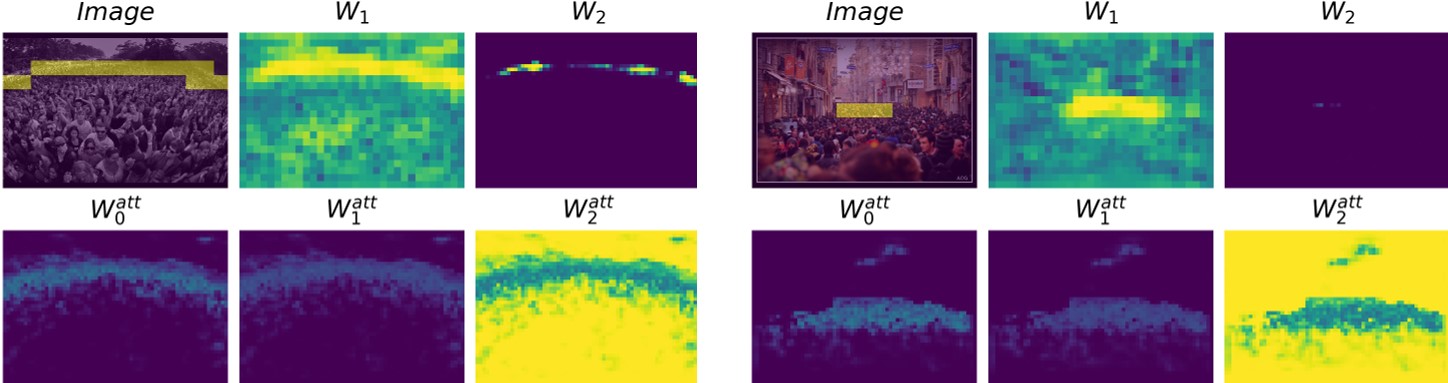}
		\end{center}
		\vspace{-15pt}
		\caption{Visualization of $W_i$ for SS-DCNet (cls) (top) and the attention baseline (bottom). The lighter the image is, the greater the values are. In the input image, count values greater than $C_{max}$ are indicated by yellow regions.}
		\label{fig:w_visual}
	\end{figure*}

	To highlight the difference between S-DC and spatial attention (SA), we remove the division decider, generate a $3$-channel output conditioned on $F_2$, then normalize it with softmax to obtain $W_0^{att}$, $W_1^{att}$ and $W_2^{att}$. The final count is merged as $W_0^{att}*upsample(C_0)+W_1^{att}*upsample(C_1)+W_2^{att}*C_2$. In SHTech PartA, SA achieves $64.1$ $MAE$ and $109.9$ $MSE$, worse than SS-DCNet. As 
	shown in
	the visualization of $W_i^{att}$ in Fig.~\ref{fig:w_visual}, we find SA only focuses on the highest resolution, and no effect of division is observed. Instead, SS-DCNet learns to divide local patches when local counts are greater than $C_{max}$. In addition, SS-DCNet executes fusion recursively, while SA fuses the prediction in a single step.
	
	\subsection{Comparison with State of the Art Methods}
	According to the ablation study, the final configurations of SS-DCNet are summarized in Table~\ref{tab:compare_setting}. 
	
	\begin{figure*}
		\begin{minipage}{\textwidth}

			\makeatletter\def\@captype{table}\makeatother
			\begin{minipage}{.50\textwidth}
				\caption{
				Performance on the test set of ShanghaiTech~\cite{MCNN_2016_CVPR} dataset. The best performance is in boldface.}
				\centering
				\footnotesize
				\vspace{-10pt}
				\begin{tabular}{l|ccc|ccc}
					\hline
					&\multicolumn{3}{c|}{Part A} &
					\multicolumn{3}{c}{Part B}\\
					\cline{2-7}
					Method & MAE &MSE&rMAE& MAE &MSE&rMAE\\
					\hline
					IG-CNN~\cite{Divide_grow_2018_CVPR}&72.5	&118.2&---	&13.6&21.1&---\\
					DRSAN~\cite{DRSAN2018Crowd}&69.3	&96.4&--- &11.1&18.2&---\\
					CSRNet~\cite{CSRNet_2018_CVPR}&68.2 & 115.0 &16.61\% &10.6&16.0&8.33\%\\
					SANet~\cite{SANet_2018_ECCV}&67.0	&104.5&--- &8.4&13.6&---\\
					SPN~\cite{SPN_2019_WACV}&61.7&99.5&--- &9.4&14.4&---\\
					BL~\cite{iccv2019bayesian}&62.8&101.8&15.19\% &7.7&12.7&5.94\%\\	
					PaDNet~\cite{tip2019pan}&59.2& 98.1&--- &8.1&12.2&---\\
					SPANet~\cite{iccv2019spanet}&59.4&92.5&---&\textbf{6.5}&\textbf{9.9}&---\\	
					PGCNet~\cite{yan2019perspective}	&57.0	&\textbf{86.0}&--- &8.8	&13.7&---\\
					\hline
					S-DCNet~\cite{xhp2019SDCNet} & 58.3 & 95.0&13.94\%  & 6.7	  & 10.7&\textbf{5.36\%}  \\
					SS-DCNet (reg) & 59.5 & 95.0&14.35\%& 7.7 & 11.1&6.96\%	  \\	
					SS-DCNet (cls) & \textbf{56.1}& 88.9&\textbf{13.78\%}& 6.6 & 10.8&5.40\%\\ 
					\hline
				\end{tabular}  
				\label{tab:compare_SHAB}
			\end{minipage}
			\hfill
			\makeatletter\def\@captype{table}\makeatother
			\begin{minipage}{.45\textwidth}
				\caption{%
				Performance on the test set of UCF\_CC\_50~\cite{UCFCC50_2013_CVPR} dataset.
				The best performance is in boldface.}
				\vspace{-10pt}
				\centering
				\footnotesize
				\addtolength{\tabcolsep}{4pt}
				\begin{tabular}{l|ccc}
					\hline
					Method & MAE &MSE&rMAE\\
					\hline
					Idrees~\etal~\cite{UCFCC50_2013_CVPR}&468.0&590.3&---\\
					Zhang~\etal~\cite{Zhang_2015_CVPR}&467.0&498.5&---\\
					IG-CNN~\cite{Divide_grow_2018_CVPR}&291.4&349.4&---\\
					D-ConvNet~\cite{DeepNegCor_2018_CVPR}&288.4&404.7&---\\
					CSRNet~\cite{CSRNet_2018_CVPR}&266.1& 397.5&30.22\%\\
					
					SANet~\cite{SANet_2018_ECCV}&258.4	&334.9&---\\
					
                    SPANet~\cite{iccv2019spanet}&232.6&311.7&---\\

					DRSAN~\cite{DRSAN2018Crowd}&219.2	&\textbf{250.2}&---\\
					
					BL~\cite{iccv2019bayesian}&213.8 &310.5&20.46\%\\
					
					PaDNet~\cite{tip2019pan}&185.8 &278.3&---\\

					\hline
					S-DCNet~\cite{xhp2019SDCNet} & 204.2 & 301.3&22.21\%\\
					SS-DCNet (reg) & 189.1 & 287.0&\textbf{19.74\%}\\
					SS-DCNet (cls) & \textbf{179.2} & 252.8&20.50\%\\
					
					\hline
				\end{tabular}
				\label{tab:compare_UCF_CC}    
			\end{minipage}
			\vfill
			\makeatletter\def\@captype{table}\makeatother
			\begin{minipage}{.3\textwidth}
				\caption{
				Performance on the test set of UCF-QNRF~\cite{Compose_Loss_2018_ECCV} dataset. ``$r$'' denotes image resizing used in~\cite{iccv2019bayesian}. The best performance is in boldface.}
				\vspace{-10pt}
				\centering
				\footnotesize
				\addtolength{\tabcolsep}{-3pt}
				\begin{tabular}{l|c|ccc}
					\hline
					Method &r& MAE &MSE&rMAE\\
					\hline
					TEDnet~\cite{cvpr2019TEDNet}& \texttimes&113&188&---\\
					CG-DRCN~\cite{iccv2019pushing}& \texttimes&112.2&176.3&---\\
					BL~\cite{iccv2019bayesian}& \checkmark&86.4&152.0&\textbf{12.24\%}\\
					PaDNet~\cite{tip2019pan}& \checkmark&96.5&170.2&---\\
					CSRNet~\cite{CSRNet_2018_CVPR}& \checkmark&98.2&157.2&16.51\%\\
					\hline
					S-DCNet~\cite{xhp2019SDCNet} & \texttimes& 104.4&176.1&17.31\%	\\
					S-DCNet~\cite{xhp2019SDCNet} & \checkmark& 97.7&167.6&14.58\%	\\
					SS-DCNet (reg) & \checkmark& 92.4&158.7&12.91\%	\\
					SS-DCNet (cls) & \checkmark& \textbf{81.9} &\textbf{143.8}&12.64\% \\
					\hline
				\end{tabular}
				\label{tab:compare_UCF-QNRF}    
			\end{minipage}
			\hfill
			\makeatletter\def\@captype{table}\makeatother
			\begin{minipage}{.4\textwidth}
				\caption{
				Performance on the test set of TRANCOS~\cite{TRANCOSdataset_IbPRIA2015} dataset. The best performance is in boldface.}
				\centering
				\footnotesize
				\addtolength{\tabcolsep}{-4pt}
				\begin{tabular}{l|cccc}
					\hline
					Method &GAME(0) &GAME(1)&GAME(2)&GAME(3)\\
					\hline
					CCNN~\cite{O2016Towards_CCNN}&12.49 &16.58 &20.02 &22.41\\
					Hydra-3s~\cite{O2016Towards_CCNN}&10.99 &13.75 &16.69 &19.32\\
					CSRNet~\cite{CSRNet_2018_CVPR}&3.56 &5.49 &8.57 &15.04\\
					SPN~\cite{SPN_2019_WACV}&3.35 &4.94 &6.47 &9.22\\
					\hline
					S-DCNet~\cite{xhp2019SDCNet} & 2.92 &4.29 &5.54&7.05\\
					SS-DCNet (reg) & 2.73 &3.82 &5.05 &6.72\\
					SS-DCNet (cls) & \textbf{2.42} & \textbf{3.30} &\textbf{4.55} & \textbf{6.17}\\
					\hline
				\end{tabular}
				\vspace{10pt}
				\label{tab:compare_Trancos}    
			\end{minipage}
			\hfill
			\makeatletter\def\@captype{table}\makeatother
			\begin{minipage}{.25\textwidth}
				\caption{
				Performance on the test set of MTC~\cite{Lu2017TasselNet} dataset. The best performance is in boldface.
				}
				\centering
				\footnotesize
				\addtolength{\tabcolsep}{-3pt}
				\begin{tabular}{l|ccc}
					\hline
					Method & MAE &MSE\\%&rMAE\\
					\hline

					DensityReg~\cite{vlaz2010denlearn}             & 11.9 & 14.8\\
					CCNN~\cite{O2016Towards_CCNN}                  & 21.0 & 25.5\\
					TasselNet~\cite{Lu2017TasselNet}		       & 6.6 & 9.6\\%&39.33\%\\
					TasselNetv2$^\dagger$~\cite{xhp2019TasselNetv2} &5.3 &9.4\\%&35.07\%\\
					CSRNet~\cite{CSRNet_2018_CVPR}&5.4&7.9\\
					\hline
					S-DCNet~\cite{xhp2019SDCNet} & 5.6 &9.1\\%&40.15\%\\
					SS-DCNet (reg) & 4.0 & 6.9\\%&29.31\%\\
					SS-DCNet (cls) & \textbf{3.9} & \textbf{6.6}\\%&20.75\% \\
					\hline
				\end{tabular}
				\label{tab:compare_MTC}    
			\end{minipage}
			\hfill
		\end{minipage}
	\end{figure*}	
	
	\subsubsection{The ShanghaiTech Dataset}
	The ShanghaiTech crowd counting dataset~\cite{MCNN_2016_CVPR} includes two parts: Part\_A and Part\_B. Part\_A has $300$ images for training and $182$ for testing. This part represents highly congested scenes. Part\_B contains $716$ images in relatively sparse scenes, where $400$ images are used for training and $316$ for testing. Quantitative results are listed in Table~\ref{tab:compare_SHAB}. The improvements of SS-DCNet are two-fold. First, with the explicit supervision of S-DC, SS-DCNet (cls) performs better than our previous S-DCNet. Second, our method outperforms the previous state-of-the-art PGCNet~\cite{yan2019perspective} in Part\_A and competitive results ($6.6$ MAE) as SPANet~\cite{iccv2019spanet}  ($6.5$ MAE) in Part\_B, respectively. These results suggest SS-DCNet is able to adapt to both sparse and crowded scenes. 
	
	\subsubsection{The UCF\_CC\_50 Dataset}
	
	UCF\_CC\_50~\cite{UCFCC50_2013_CVPR} is a tiny crowd counting dataset with $50$ images in extremely crowded scenes. The number of people within an images varies from $96$ to $4633$. We follow the 5-fold cross-validation as in~\cite{UCFCC50_2013_CVPR}. Results are shown in Table~\ref{tab:compare_UCF_CC}. Our method surpasses S-DCNet and the previous best method, PaDNet~\cite{tip2019pan}, with $12.2\%$ and $3.4\%$ relative improvements in \textit{MAE}, respectively.%
	
	\subsubsection{The UCF-QNRF Dataset} 
	UCF-QNRF~\cite{Compose_Loss_2018_ECCV} is a relatively large crowd counting dataset with $1535$ high-resolution images and $1.25$ million head annotations. There are $1201$ training images and $334$ test images. It contains extremely congested scenes where the maximum count of an image can reach $12865$. Some images in the UCF-QNRF dataset are too large, with the longer side equals to $10000$, to process the whole image. There are two ways to solve this problem: $i)$ cropping the original image into $224\times224$ sub-images following~\cite{Compose_Loss_2018_ECCV};  $ii)$ resizing the original image to make the longer side no larger than $1920$ as in~\cite{iccv2019learn_to_scale,iccv2019bayesian}, then $9$ sub-images of $\frac{1}{4}$ resolution are cropped from the original image for data augmentation as described in Section~\ref{subsec:data_pre_process}.
	Results are reported in Table~\ref{tab:compare_UCF-QNRF}. We can make following observations:
	\squishlist
	\item For S-DCNet, it works better with strategy $ii)$ than $i)$. This means that resizing is a better choice than cropping. We think the reasons are two-fold. First, the receptive field of a CNN is limited, thus it cannot cover over-size images. Second, if cropping over-size images into $224\times 224$ sub-images, the surrounding pixels of sub-images, termed `local visual context', are invisible to the CNN. However, the local visual context can provide support information to distinguish overlapped objects as demonstrated in~\cite{xhp2019TasselNetv2}, and CNNs tend to perform poorly when local context is lost.  
	\item With the explicit supervision of S-DC, SS-DCNet (cls) brings a significant improvement over S-DCNet by $15.8$ in MAE, and SS-DCNet (reg) shows by $5.3$ in MAE.
	\item SS-DCNet (reg) reports competitive results against the current state-of-the-art $BL$~\cite{iccv2019bayesian}, while SS-DCNet (cls) outperforms $BL$ by $6.8$ in MAE and $11$ in MSE.   
	\item It is worth noting that, SS-DCNet only learns from a closed set with $C_{max}=8.0$, which is only $6\%$ of the maximum count $131.5$ according to Table~\ref{tab:compare_setting}. SS-DCNet, however, generalizes to large counts effectively and predicts accurate counts. 
	\squishend 

	\subsubsection{The TRANCOS Dataset}
	Aside from crowd counting, we also evaluate SS-DCNet on a vehicle counting dataset, TRANCOS~\cite{TRANCOSdataset_IbPRIA2015}, to demonstrate the generality of SS-DCNet. TRANCOS contains $1244$ images of congested traffic scenes in various perspectives. It adopts the Grid Average Mean Absolute Error (GAME)~\cite{TRANCOSdataset_IbPRIA2015} as the evaluation metric. $GAME(L)$ divides an image into $2^L\times2^L$ non-overlapping sub-regions and accumulates of the $MAE$ over sub-regions. Larger $L$ implies more accurate local predictions. In particular, $GAME(0)$ downgrades to $MAE$. 	
	The GAME is defined by
	\begin{equation}\label{Game_loss}
	GAME(L) = \frac{1}{N}\sum\limits_{n=1}^{N}(\sum\limits_{l=1}^{4^L}|C_{pre}^l-C_{gt}^l|)\,,
	\end{equation}
	where $N$ denotes the number of images. $C_{pre}^l$ and $C_{gt}^l$ are the predicted and ground-truth count of the $L$-th sub-region, respectively. 
	Results are listed in Table~\ref{tab:compare_Trancos}. SS-DCNet surpasses other methods under all $GAME(L)$ metrics, and particularly, delivers a $33.1\%$ relative improvement than SPN~\cite{SPN_2019_WACV} on $GAME(3)$. This suggests SS-DCNet not only achieves accurate global predictions but also behaves well in local regions. 
	
	\subsubsection{The MTC Dataset}
	We further evaluate our method on a plant counting dataset, i.e., the MTC dataset~\cite{Lu2017TasselNet}. The MTC dataset contains $361$ high-resolution images of maize tassels collected from 2010 to 2015 in the wild field. In contrast to 
	pedestrians or vehicles that have similar physical sizes, maize tassels are with heterogeneous physical sizes and are self-changing over time. We believe that 
	this dataset is suitable for justifying the robustness of SS-DCNet to object-size variations. We follow the same setting as in~\cite{Lu2017TasselNet} and report quantitative results in Table~\ref{tab:compare_MTC}. 
	Although the previous best method, TasselNetv2$^\dagger$~\cite{xhp2019TasselNetv2}, already exhibits accurate results, SS-DCNet still shows a substantial degree of improvement ($26.4\%$ on $MAE$ and $29.8\%$ on MSE).
	
\begin{figure*}[!t]
		\begin{center}
			\includegraphics[width=1.00\linewidth]{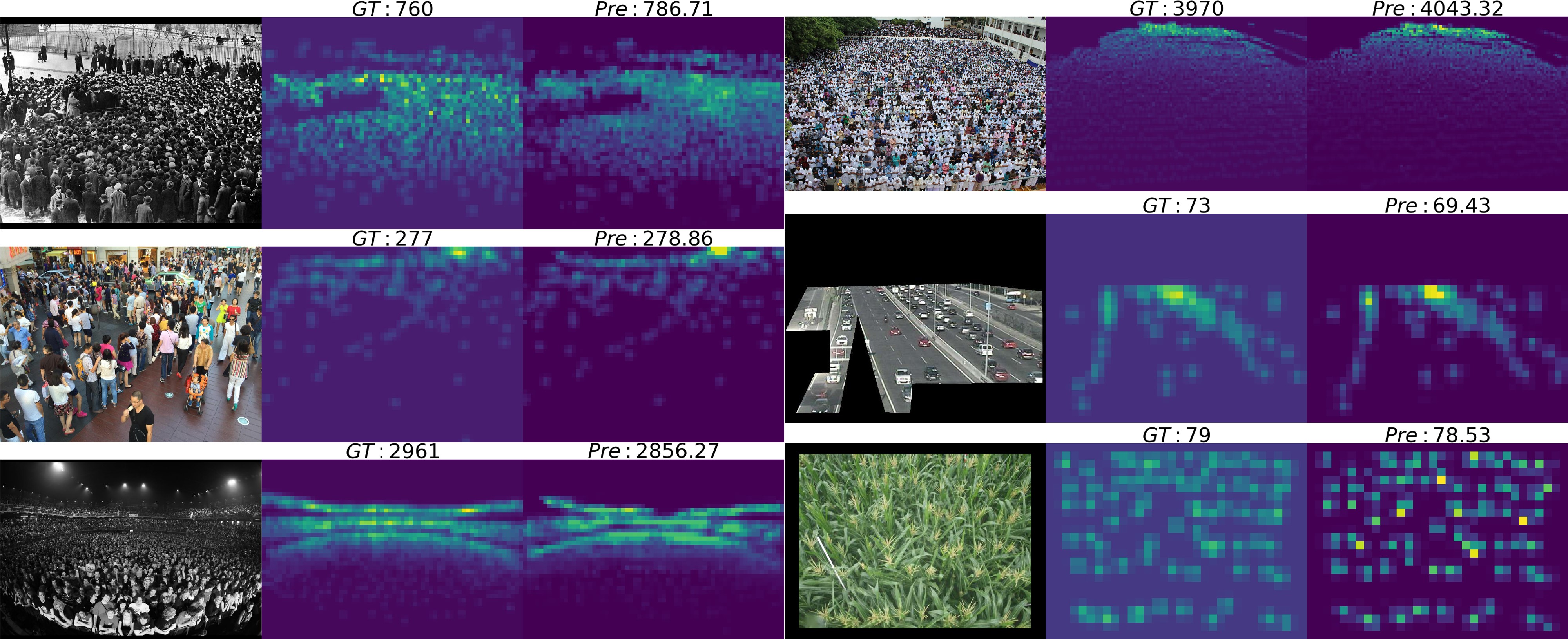}
		\end{center}
		\vspace{-15pt}
		\caption{Visualization of count map for SS-DCNet on various datasets. From left to right, for each visualization, the original image, the ground-truth count map, and the inferred count map, respectively. From top to down, left to right, the visualization of ShanghaiTech Part\_A, Part\_B~\cite{MCNN_2016_CVPR}, UCF\_CC\_50~\cite{UCFCC50_2013_CVPR}, UCF-QNRF~\cite{Compose_Loss_2018_ECCV}, TRANCOS~\cite{TRANCOSdataset_IbPRIA2015}, and MTC~\cite{Lu2017TasselNet}, respectively. The original image is zero-padded to be divisible by $64$.}
		\label{fig:6dataset_countmap_visual}
	\end{figure*}

	Qualitative results are shown in Fig~\ref{fig:6dataset_countmap_visual}.
	We can observe that SS-DCNet produces accurate predictions for various objects from sparse to dense scenes.
	
	\begin{table*}[!t] %
		\caption{Cross-dataset performance (MAE/MSE/rMAE) on the ShanghaiTech (A and B) and UCF-QNRF (QNRF) Datasets. Best performance is in boldface.}
		\centering
		\renewcommand\arraystretch{0.6}
		\setlength{\tabcolsep}{1.2mm} {
			\begin{tabular}{l|c|c|c|c|c|c }
				\hline
			Method&A $\rightarrow$ B& A$\rightarrow$QNRF& B$\rightarrow$A& B$\rightarrow$QNRF& QNRF$\rightarrow$A& QNRF$\rightarrow$B\\
            \hline
				D-ConvNet~\cite{DeepNegCor_2018_CVPR}                &49.1/ 99.2/---  &---/---/---  &140.4 /226.1/---&---/---/---&---/---/---&---/---/---\\
				SPN+L2SM~\cite{iccv2019learn_to_scale}  &\textbf{21.2}/38.7/---& 227.2/405.2/---&\textbf{126.8}/\textbf{203.9}/---&---/---/---                   &73.4/119.4/---&---/---/---\\
				BL~\cite{iccv2019bayesian}     & 16.3/30.3/\textbf{12.24\%}  & 141.6/252.4/23.41\%& 137.0/228.9/30.32\%& 208.9/41.4/25.97\%&        69.8/123.8/14.85\%&15.3/26.5/10.99\%\\

				\hline	
				regression&23.6/\textbf{35.0}/18.91\% &172.7/320.6/19.66\% &133.9/228.4/29.20\% &230.3/419.3/27.09\% &71.7/116.9/16.34\%&14.2/23.3/11.31\%\\
				classification&21.4/36.6/16.49\%&173.7/323.1/21.64\%&179.4/313.4/31.60\%&281.7/512.3/28.49\% &134.7/259.5/22.81\%&21.9/47.8/14.24\%\\
				
				SS-DCNet (reg)&22.9/35.2/19.55\%&160.6/299.3/19.27\% &137.1/235.8/30.77\%&222.1/399.5/27.98\% &69.0	  /115.8/15.53\%&12.1/\textbf{20.8}/9.20\%		\\ 		
				SS-DCNet (cls)&\textbf{21.2}/39.5/15.89\% &\textbf{151.8}/\textbf{270.4}/\textbf{18.36\%} &130.0/209.6/\textbf{27.82\%} &\textbf{166.5}/\textbf{281.8}/\textbf{23.14\%} &\textbf{61.8}/\textbf{102.8}/\textbf{13.79\%}&\textbf{11.8}/21.8/\textbf{7.98\%}\\	
				\hline
				
			\end{tabular}
		}
		\label{tab:Cross experiments}
	\end{table*}
	
	\begin{figure}[!t]
		\begin{center}
			\includegraphics[width=1.0\linewidth]{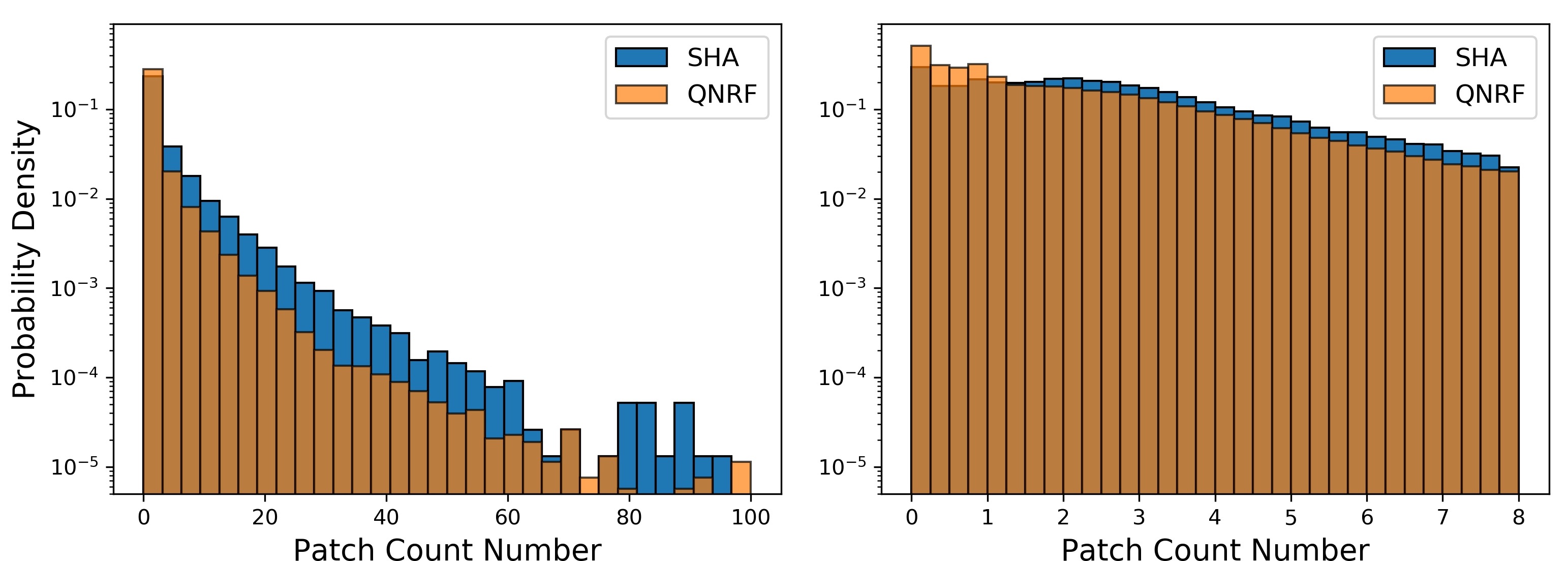}
		\end{center}
		\vspace{-15pt}
		\caption{ The probability density histograms of patch count numbers of SHA and QNRF. The area of each histogram adds up to $1$. The left histogram shows the probability distribution without S-DC, and the right the probability distribution after S-DC, where count values are transformed into the closed-set $[0,8]$. }
		\label{fig:SHA_QNRF_distribute}
	\end{figure}
	
	\subsection{Cross-Dataset %
	Evaluation
	}
	We further conduct cross-dataset experiments on the ShanghaiTech~\cite{MCNN_2016_CVPR} (A and B) and UCF-QNRF (QNRF)~\cite{Compose_Loss_2018_ECCV} datasets to show the %
	generalization 
	ability of SS-DCNet. Quantitative results are shown in Table~\ref{tab:Cross experiments}. The `regression' and `classification' methods are baselines for SS-DCNet (reg) and S-DCNet (cls), respectively, which adopt the VGG16~\cite{Simonyan2014Very_VGG16} as the feature encoder and R-Counter/C-Counter in SS-DCNet but do not apply S-DC. We can make following observations:
	\squishlist  
	\item Consistent improvements in MAE are observed when comparing SS-DCNet (reg/cls) to its baselines. Especially in SS-DCNet (cls) vs.\ baseline cls, the $MAE$ of \mbox{$B\rightarrow QNRF$} shows a $40.89\%$ relative improvement when S-DC is added. 
	\item Two types of SS-DCNet report superior or at least competitive results than other state-of-the-art methods under all cross-dataset tasks, which suggest SS-DCNet has strong transferring ability. 
	\item SS-DCNet (cls) transferred from the QNRF dataset reports even better results ($61.8$ MAE) than most state-of-the-arts methods (e.g., $68.2$ MAE for CSRNet and $67.0$ MAE for SANet) trained on the ShanghaiTech dataset. 
	\item All methods trained on the ShanghaiTech~\cite{MCNN_2016_CVPR} dataset report worse cross-dataset results than trained directly on the target datasets. By contrast, all methods trained on the QNRF~\cite{Compose_Loss_2018_ECCV} dataset exhibits at least competitive transferring results against state-of-the-art methods trained on the target dataset. This may be attributed to the fact that the ShanghaiTech dataset is too small, with only $300$ training samples in the Part\_A and $400$ in the Part\_B, to train a robust model, while 
	the QNRF dataset provides sufficient training samples.
	\squishend

	Overall, \textit{SS-DCNet demonstrates state-of-the-art results in all cross-dataset experiments}. The good performance of SS-DCNet may be explained from its implicit transferring ability in the output space, which shares the same spirit with~\cite{AdaptOutput2018CVPR}. To justify this, we analyze the case of $QNRF\rightarrow A$ and visualize the distribution of count values with and without S-DC in Fig.~\ref{fig:SHA_QNRF_distribute}. It can be observed that, the distribution of count values varies significantly between SHA and QNRF without S-DC, but the divergence of the distribution narrows down after count values are transformed into a closed set $[0,8]$. We further compute the Jensen–Shannon divergence~\cite{osterreicher2003JS} $J_s$ to quantify the divergence of the distribution, and find that $J_s=0.0178$ between SHA and QNRF without S-DC and $J_s = 0.0133$ with S-DC. The smaller $J_s$ is, the smaller divergence between two distributions shows. This means the divergence of the output (count value) space reduces after closed-set transformation. We believe this is the main reason why SS-DCNet reports remarkable performance on the task of $QNRF\rightarrow A$. 

	\section{Conclusion}
	Counting is an open-set problem in theory, but only a finite closed set of training data can be observed in reality. This is particularly true because any dataset is always a sampling of the real world. Inspired by the
	decomposition 
	property of counting, we have proposed to transform the open-set counting into a closed-set problem, and implement this transformation with the idea of S-DC. We propose supervised S-DC in a deep counting network,  termed SS-DCNet. We provide a theoretical analysis showing why the transformation from the open set to closed set makes sense. 
	Experiments on both synthetic data and real
	benchmark datasets
	show that, even given a closed training set, SS-DCNet can effectively generalize to open-set scenarios. Furthermore, SS-DCNet shows its good generalization ability via %
	cross-dataset performance.
	
	Many vision tasks are open-set by nature, depth estimation for example, 
	while it is not immediately clear on how to tranform them into a closed set like counting.
	It would be interesting to explore 
	how to transform other vision tasks into a closed set setting.

	\bibliographystyle{ieee_fullname.bst}
	\bibliography{egbib}

	\begin{IEEEbiography}[{\includegraphics[width=1in,height=1.25in,clip,keepaspectratio]{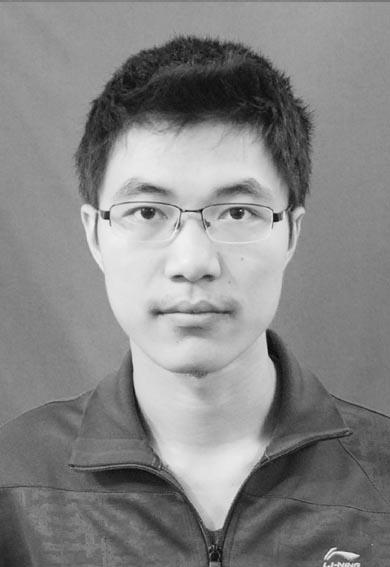}}]{Haipeng Xiong}
		received the B.S. degree from Huazhong University of science and Technology, Wuhan, China, in 2018. He is currently pursuing the M.S. degree with the School of Artificial Intelligence and Automation, Huazhong University of Science and Technology, Wuhan, China. 
		
		He currently researches object counting and its applications in agriculture. His research interests include math, machine learning and computer vision.
	\end{IEEEbiography}
	\vskip 0pt plus -1fil
	\begin{IEEEbiography}[{\includegraphics[width=1in,height=1.25in,clip,keepaspectratio]{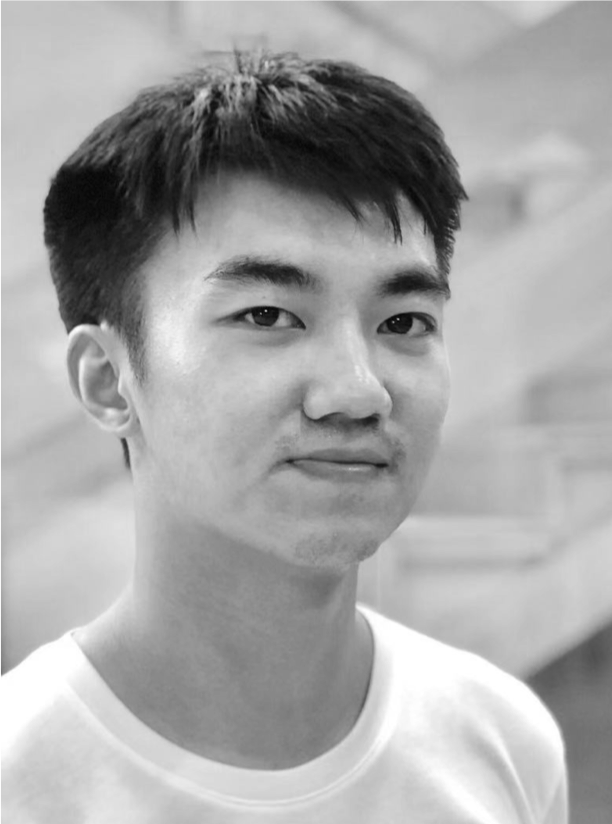}}]{Hao Lu}
		received the Ph.D. degree from Huazhong University of science and Technology, Wuhan, China, in 2018.
		
		He is currently a Postdoctoral Fellow with the School of Computer Science, the University of Adelaide. His research interests include computer vision, image processing and machine learning. He has worked on topics including visual domain adaptation, fine-grained visual categorization, as well as miscellaneous computer vision applications in agriculture. His current interests are object counting and dense prediction problems.
	\end{IEEEbiography}
	\vskip 0pt plus -1fil
	\begin{IEEEbiography}[{\includegraphics[width=1in,height=1.25in,clip,keepaspectratio]{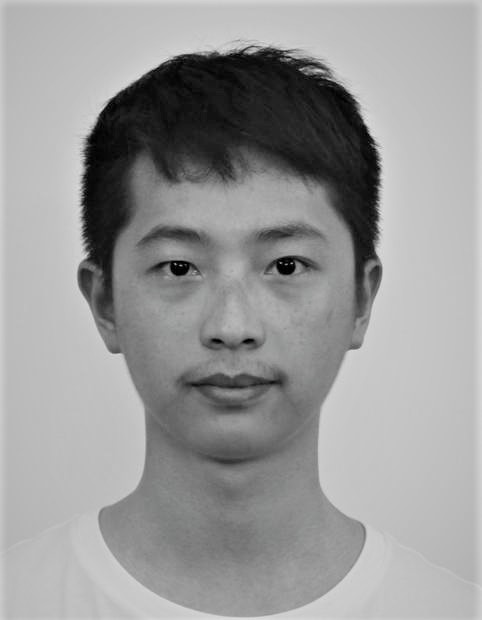}}]{Chengxin Liu}
		received the B.S. degree from Huazhong University of science and Technology, Wuhan, China, in 2018. He is currently pursuing the M.S. degree with the School of Artificial Intelligence and Automation, Huazhong University of Science and Technology, Wuhan, China. He currently researches object tracking. 
	\end{IEEEbiography}

	\vskip 0pt plus -1fil
	\begin{IEEEbiography}[{\includegraphics[width=1in,height=1.25in,clip,keepaspectratio]{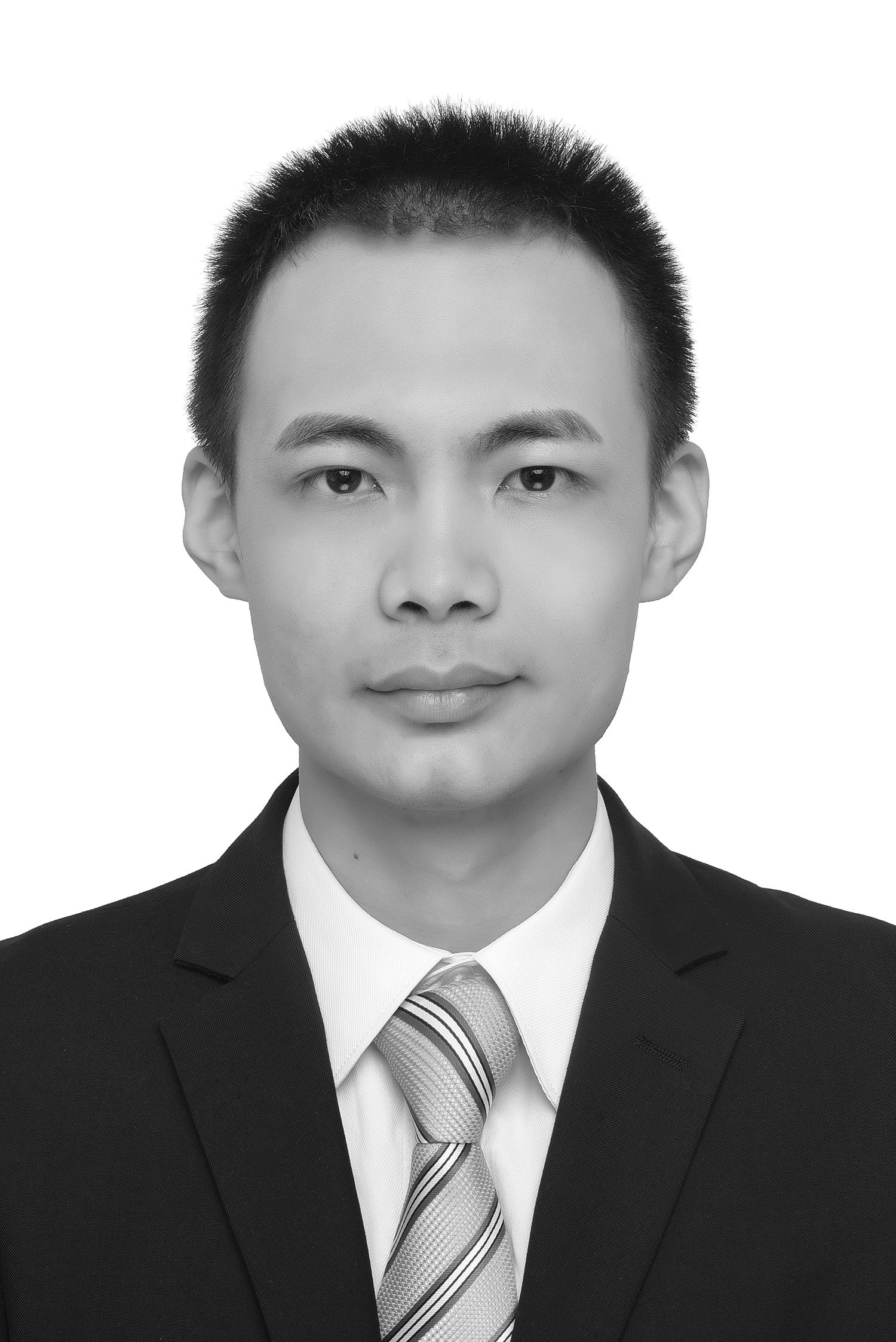}}]{Liang Liu}
		received the B.S. degree from Huazhong University of science and Technology, Wuhan, China, in 2016. He is currently pursuing the Ph.D. degree with the School of Artificial Intelligence and Automation, Huazhong University of Science and Technology, Wuhan, China.
		
		His research interests include computer vision and machine learning, with particular emphasis on object counting and various computer vision applications in agriculture.
	\end{IEEEbiography}

	\vskip 0pt plus -1fil
	\begin{IEEEbiographynophoto}
	{Chunhua Shen}
		is a Professor of Computer Science, at the University of Adelaide, Australia.
	\end{IEEEbiographynophoto}
	
	\vskip 0pt plus -1fil
	\begin{IEEEbiography}[{\includegraphics[width=1in,height=1.25in,clip,keepaspectratio]{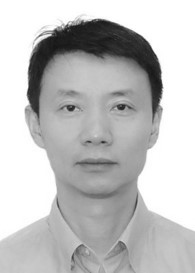}}]{Zhiguo Cao}
		received the B.S. and M.S. degrees in communication and information system from the University of Electronic Science and Technology of China, Chengdu, China, and the Ph.D. degree in pattern recognition and intelligent system from Huazhong University of Science and Technology, Wuhan, China.
		
		He is currently a Professor with the School of Artificial Intelligence and Automation, Huazhong University of Science and Technology. He has authored dozens of papers at international journals and conferences, which have been applied to automatic observation system for object recognition in video surveillance system, for crop growth in agriculture and for weather phenomenon in meteorology based on computer vision. His research interests spread across image understanding and analysis, depth information extraction and object detection.
		
		Dr. Cao's projects have received provincial or ministerial level awards of Science and Technology Progress in China.
	\end{IEEEbiography}	
	\vfill
	
\end{document}